\documentclass[journal]{IEEEtran}
\usepackage[T1]{fontenc}
\usepackage{graphicx}
\usepackage{times}
\usepackage{helvet}
\usepackage{courier}
\usepackage{amsmath}
\usepackage{algorithm}
\usepackage{algorithmic}
\usepackage{csquotes} 
\usepackage{color}
\usepackage{paralist}
\usepackage{amssymb}
\usepackage{indentfirst}
\usepackage{subfigure}
\usepackage{float}
\usepackage{multirow}
\usepackage{cite}
\usepackage{mathrsfs}

\usepackage{mathrsfs} 
\usepackage{amsfonts}
\usepackage{todonotes}
\usepackage{pgfplots} 
\usepackage{epstopdf}
\usepackage{epsfig}
\pgfplotsset{compat=newest}
\usepackage{color}
\usepackage{tabularx}

\definecolor{forestgreen}{RGB}{0,139,69}

\usepackage{xcolor}
\definecolor{citecolor}{HTML}{0071bc}
\usepackage[colorlinks, linkcolor=red,  anchorcolor=blue, citecolor=citecolor]{hyperref} 

\usepackage{xcolor}
\definecolor{SeaGreen4}{RGB}{0,205,102} 
\definecolor{SlateBlue}{RGB}{106,90,205} 
\definecolor{DarkRed}{RGB}{178,34,34}

\usepackage{textcomp,booktabs}
\usepackage{amssymb}
\usepackage{pifont}

\usepackage{makecell}

\usepackage{colortbl}
\definecolor{mygray}{gray}{.9}
\definecolor{mypink}{rgb}{.99,.91,.95}
\definecolor{mycyan}{cmyk}{.3,0,0,0}

\usepackage[switch]{lineno}

\begin{document}

\title{Vehicle-centric Perception via Multimodal Structured Pre-training}

\author{Wentao Wu, Xiao Wang*, \emph{Member, IEEE}, Chenglong Li*, Jin Tang, Bin Luo, \emph{Senior Member, IEEE}

\thanks{ $\bullet$ Wentao Wu, Chenglong Li are with Information Materials and Intelligent Sensing Laboratory of Anhui Province, Anhui Provincial Key Laboratory of Multimodal Cognitive Computation, the School of Artificial Intelligence, Anhui University, Hefei 230601, China. 
(email: wuwentao0708@163.com, lcl1314@foxmail.com)} 

\thanks{ $\bullet$  Xiao Wang, Jin Tang, Bin Luo are with the School of Computer Science and Technology, Anhui University, Institute of Artificial Intelligence, Hefei Comprehensive National Science Center, Hefei 230601, China. 
(email: xiaowang@ahu.edu.cn) 
}

\thanks{* Corresponding author: Xiao Wang and Chenglong Li} 
}

\markboth{IEEE Transactions on Circuits and Systems for Video Technology, 2025}   
{Shell \MakeLowercase{\textit{et al.}}: Bare Demo of IEEEtran.cls for IEEE Journals}

\maketitle

\begin{abstract}
Vehicle-centric perception plays a crucial role in many intelligent systems, including large-scale surveillance systems, intelligent transportation, and autonomous driving. 
Existing approaches typically employ general pre-trained weights to initialize backbone networks, followed by task-specific fine-tuning. However, these models lack effective learning of vehicle-related knowledge during pre-training, resulting in poor capability for modeling general vehicle perception representations. To handle this problem, we propose VehicleMAE-V2, a novel vehicle-centric pre-trained large model. By exploring and exploiting vehicle-related multimodal structured priors to guide the masked token reconstruction process, our approach can significantly enhance the model's capability to learn generalizable representations for vehicle-centric perception. Specifically, we design the Symmetry-guided Mask Module (SMM), Contour-guided Representation Module (CRM) and Semantics-guided Representation Module (SRM) to incorporate three kinds of structured priors into token reconstruction including symmetry, contour and semantics of vehicles respectively. SMM utilizes the vehicle symmetry constraints to avoid retaining symmetric patches and can thus select high-quality masked image patches and reduce information redundancy. CRM minimizes the probability distribution divergence between contour features and reconstructed features and can thus preserve holistic vehicle structure information during pixel-level reconstruction. SRM aligns image-text features through contrastive learning and cross-modal distillation to address the feature confusion caused by insufficient semantic understanding during masked reconstruction. To support the pre-training of VehicleMAE-V2, we construct Autobot4M, a large-scale dataset comprising approximately 4 million vehicle images and 12,693 text descriptions. Extensive experiments on five downstream tasks demonstrate the superior performance of VehicleMAE-V2.
The source code, dataset, and pre-trained large models are available on 
\url{https://github.com/Vehicle-AHU/VehicleMAE}.
\end{abstract}

\begin{IEEEkeywords}
Pre-trained Vision Model; Masked Auto-Encoder; Vehicle-centric Perception; Vision and Language; Self-Supervised Learning 
\end{IEEEkeywords}

\IEEEpeerreviewmaketitle

\section{Introduction} \label{sec1}

\IEEEPARstart{W}{ith} the rapid development of artificial intelligence, more and more vehicle-centric perception problems have been widely discussed. For vehicle perception in autonomous driving and smart cities, core computer vision tasks include vehicle classification~\cite{yu2022embedding}, detection~\cite{liang2024scene}, segmentation~\cite{wang2024class}, re-identification~\cite{ran2025context}, and attribute recognition~\cite{wang2024pedestrian}. In real-world scenarios, these tasks face a number of challenges, including low illumination, motion blur, and occlusion, to name a few. To address these challenges, current researchers usually study these research topics in a relatively independent way in the early stage of the third wave of deep learning techniques. The popular paradigm is that a small dataset is collected and annotated for each topic to train a deep neural network from scratch or based on pre-trained backbone networks like ResNet~\cite{he2016deep} or ViT~\cite{VIT}. While this paradigm has achieved some progress, the intricate nature of real-world environments means these challenges are yet to be fully resolved.

\begin{figure}
\centering
\includegraphics[width=3in]{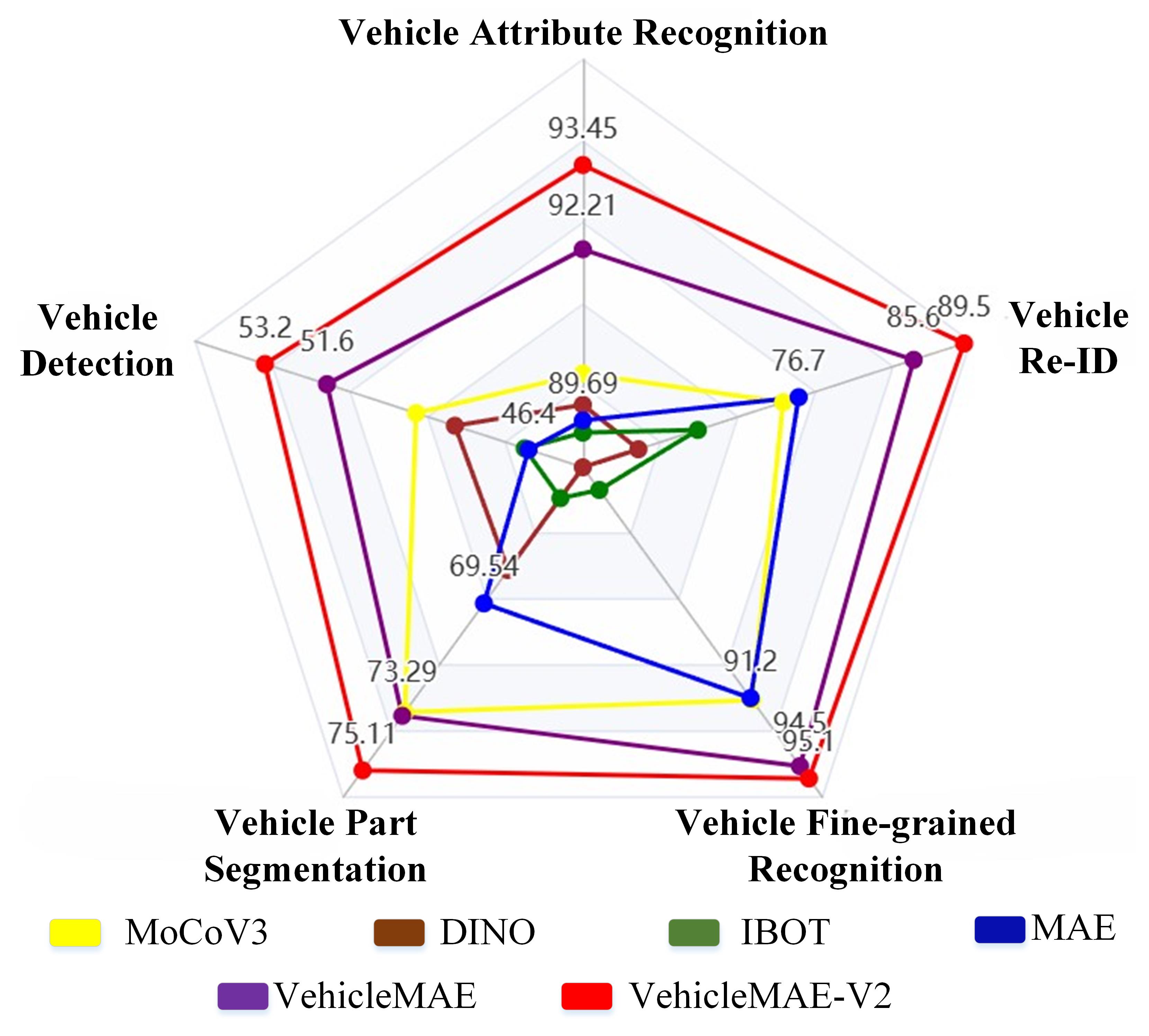}
\caption{Comparison between existing large models and our newly proposed VehicleMAE-V2 on five downstream vehicle related tasks.}
\label{fig:radarmap} 
\end{figure}

\begin{figure*}
\centering
\includegraphics[width=0.9\textwidth]{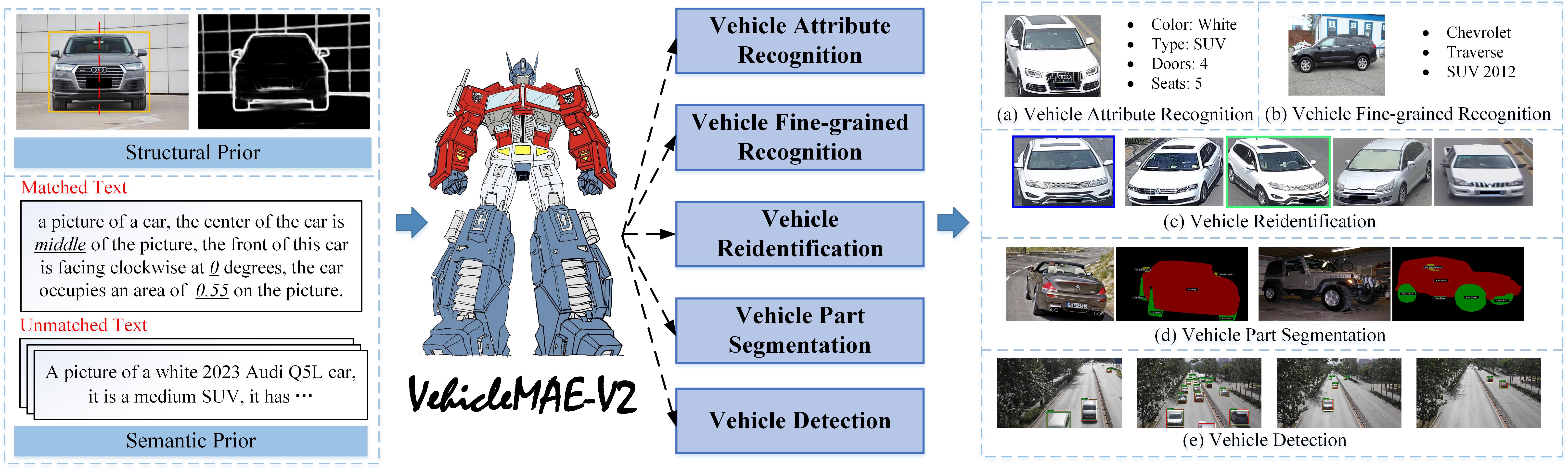}
\caption{
An illustration of our newly proposed pre-trained vehicle-centric large model VehicleMAE-V2, which is pre-trained on 4M vehicle images based on structural and semantic priors. 
Five downstream tasks are evaluated to validate the effectiveness and generalization of the proposed VehicleMAE-V2.
}
\label{fig:firstIMG} 
\end{figure*}

Recently, the pre-trained large models~\cite{yuan2024surveillance, chen2023beyond} have garnered increasing attention within the computer vision community, initially making a splash in natural language processing. For example, the GPT series~\cite{radford2018GPT1, radford2019GPT2, brown2020GPT3, openai2023gpt4} and Llama series~\cite{touvron2023llama} demonstrate their powerful performance on different tasks, e.g., translation, reasoning and representation learning. The representative vision Transformer networks ViT~\cite{VIT} and Swin-Transformer~\cite{liu2021swintransformer} achieve compelling results on various vision tasks, such as object detection and tracking, segmentation and recognition. Usually, these large models are pre-trained in a self-supervised learning manner on the large-scale pre-training corpus. There are also many multi-modal pre-trained large models~\cite{li2023blip, xue2024ulip} which further extend the scope of input modalities into language, audio, LiDAR, event/spike stream, etc.

Among them, the Masked Auto-Encoder (MAE)~\cite{he2022mae} is a representative pre-training framework that masks a high proportion of the input images and reconstructs the masked regions for visual representation learning. Although pre-training foundational vision models on large-scale vehicle images is intuitively feasible, the lack of learning mechanisms tailored to vehicle characteristics can limit the model's ability to learn generalizable representations for vehicle perception tasks. Considering that a vehicle is actually a special target object with contours consisting of curves or straight lines, unified colors, and bilateral symmetry. In addition, the manufacturer provides detailed parameter documentation and other information for each vehicle, and we can actually obtain a considerable amount of multimodal data for pre-training vehicle models. Based on the above observations and reflections, this paper presents a large vehicle-centric model pre-trained based on multimodal structured priors enhanced MAE framework, termed VehicleMAE-V2. As a general pre-trained large model, it can be adopted in various vehicle-related downstream tasks, such as attribute recognition, fine-grained recognition, re-identification, partial segmentation, and vehicle detection, as shown in Figure~\ref{fig:radarmap} and Figure~\ref{fig:firstIMG}.

Specifically, our framework consists of three main modules: the Symmetry-guided Mask Module (SMM), the Contour-guided Representation Module (CRM), and the Semantics-guided Representation Module (SRM). Given the input vehicle images, we utilize a vehicle angle detector to pre-acquire the bounding box and angle information of each vehicle. The SMM utilizes this information to prioritize the selection of image patches from target regions for masking, while avoiding information redundancy caused by symmetrical patches. The unmasked image patches are fed into an asymmetric Transformer encoder-decoder for masked patch reconstruction. 
In the CRM, we extract contour maps of the input images using an edge detector and employ a shared Transformer encoder to obtain contour features. By minimizing the probability distribution divergence between contour features and reconstructed features, we supervise the reconstruction of the vehicle’s overall structural information at the pixel level. 
In the SRM, we design two types of textual descriptions. The first is unpaired textual descriptions related to vehicle attributes. For this, we introduce a pre-trained CLIP~\cite{radford2021learning} visual-language encoder to obtain visual and textual representations. By constructing a cross-modal distillation loss between CLIP-encoded features and the reconstructed features output by the decoder, we learn rich vehicle knowledge from the textual descriptions. The second is paired textual descriptions relevant to the image content. We use the CLIP text encoder to extract text embeddings and perform contrastive learning with the visual features extracted by the encoder, thereby enhancing the model's semantic understanding capability. These designs collectively contribute to achieving efficient and high-performance vehicle pre-training.
An overview of the VehicleMAE-V2 can be found in Figure~\ref{fig:framework}. 

To pre-train our VehicleMAE-V2 effectively, we collect a new large-scale dataset, which contains approximately 4 million vehicle images, named Autobot4M. The sources of these images, including existing datasets, public visual surveillance systems, and vehicle websites, fully reflect the key challenges in vehicle-centric perception. For example, illumination variation, motion blur, viewpoints, and occlusions are all considered. It is worth noting that part of these images is crawled from the Internet, and the corresponding language descriptions are also used for multimodal pre-training in our framework. Please check Section~\ref{sec4} for a better understanding of the pre-training dataset.

To sum up, the key contributions of this paper can be summarized as the following three aspects:

$\bullet$ We propose a novel vehicle-centric perceptual multimodal pre-training framework, termed VehicleMAE-V2. By integrating vehicle-related multimodal structured information during pre-training, it significantly enhances the performance of downstream vehicle perception tasks.

$\bullet$ We design a set of prior-guided modules that integrate symmetry, contour, and semantics priors into the masked token reconstruction process, thereby enhancing the model's ability to learn generalizable representations for vehicle perception.

$\bullet$ We construct a large-scale vehicle pre-training dataset, termed Autobot4M, which comprises 4 million images and over 10,000 textual descriptions of various vehicle models, covering diverse scenarios and challenging conditions.

This paper is an extension of our previous work VehicleMAE~\cite{wang2024structural}, which was published at the international conference AAAI 2024. The key improvements can be summarized as follows: 
\textit{1) Larger Pre-training Dataset:} We significantly expanded the Autobot1M dataset from the conference version to Autobot4M, which is four times larger, containing 4 million samples. This provides substantial support for large-scale model pre-training. During dataset expansion, we specifically focused on addressing the lack of in-vehicle perspective images in Autobot1M, supplementing 1 million such vehicle images in Autobot4M. Additionally, images of complex scenarios such as nighttime and rainy conditions were significantly expanded in Autobot4M. This dataset will serve as a valuable resource for other researchers in the field of vehicle-centric studies.  
\textit{2) Symmetry-guided Mask Module:} In the conference version, we adopted the same random masking strategy as MAE. However, this approach of randomly selecting masked patches tends to result in a large number of non-critical regions being masked. Additionally, since vehicles are rigid objects with obvious left-right symmetry, symmetric image patches contain redundant information, leading to inefficiency when fed into the encoder. To address this, we designed a Symmetry-guided Mask Module for VehicleMAE-V2. By leveraging angle and bounding box information from the detector, we compute the symmetry axis of the target to avoid retaining symmetric patches in the unmasked regions. Furthermore, since detectors may occasionally fail, our module integrates three masking strategies: random masking, box-guided masking, and symmetry-guided masking.
\textit{3) Semantics-guided Representation Module:}  In the conference version, we collected over 10,000 vehicle model description texts. However, these texts were not strictly paired with images and lacked descriptive content about the visual features. To address this, we leveraged the angle and bounding box information from the angle detector to generate paired descriptive texts for each image (images without detectable angles or bounding boxes were discarded). Through vision-language contrastive learning, we enhanced the model's ability to perceive semantic information in images.
\textit{4) Diverse Downstream Tasks:} In the conference version, our pre-training dataset consisted of small-scene images of individual vehicles, which were not well-suited for large-scene detection tasks. Therefore, the downstream tasks were only focused on small-scene images. In subsequent research, we proposed a novel framework called VFM-Det~\cite{wu2024vfm}, which applies VehicleMAE to object detection. As a result, this study expands the validation to five vehicle-related downstream tasks by adding a vehicle detection task. Additionally, more comprehensive experiments were conducted to further verify the effectiveness of the proposed vehicle pre-training strategy.

\section{Related Works} \label{sec2}


\subsection{Pre-trained Large Models} 
In recent years, pre-training models on large-scale datasets through self-/un-supervised learning have been a hot research topic. Self-supervised learning typically involves training on unlabeled data using automatically generated labels or tasks. Common pre-training models can be categorized into two types based on different proxy tasks: \textit{contrastive learning-based} and \textit{reconstruction-based} pre-training methods. 
Specifically, contrastive learning–based methods aim to pull similar samples closer while pushing dissimilar samples apart, thereby learning a representation space that facilitates discrimination among different data categories. In early computer vision studies, contrastive learning commonly relies on paired samples constructed through data augmentation. For example, MoCo~\cite{he2020momentum} introduces momentum encoders to reduce the computational cost induced by large numbers of negative samples, while SimCLR~\cite{chen2020simple} enhances contrastive learning by adopting richer data augmentations and larger batch sizes. Building upon this paradigm, recent studies further extend contrastive learning to multimodal and finer-grained representation learning scenarios to alleviate representation discrepancies across different modalities or domains. For instance, CDC~\cite{qiu2024camera} proposes a camera-aware differentiated clustering and focal contrastive learning method to reduce pseudo-label noise for unsupervised vehicle re-identification; Wang et al.~\cite{wang2025contrastive} propose a contrastive learning framework based on class-level and cluster-level memory dictionaries, jointly pre-training on source and target domains to facilitate more accurate pseudo-label generation during fine-tuning;
SDCluster~\cite{xu2025sdcluster} introduces pixel-level clustering constraints under the contrastive learning paradigm, removing ineffective prototypes and preserving critical semantic information through prototype and semantic consistency constraints.

Another mainstream category comprises reconstruction-based pre-training methods, which perform self-supervised learning by reconstructing masked portions of the input within the model, thereby encouraging the learning of expressive feature representations. This paradigm is widely adopted in both natural language processing and computer vision. For example, BERT~\cite{kenton2019bert} conducts pre-training via masked language modeling on large-scale corpora, while MAE~\cite{he2022mae} and BEiT~\cite{baobeit} learn visual representations by predicting the original pixels of masked image patches. In recent years, reconstruction strategies further extend to various domain-specific applications. In the 3D perception domain, Zhu et al.~\cite{zhu2025driving} propose a unified framework that mutually enhances 3D reconstruction and image inpainting for high-quality 3D face reconstruction; Gong et al.~\cite{gong2025rethinking} combine contrastive learning with masked reconstruction to learn more discriminative 3D action representations. In the autonomous driving domain, Ni et al.~\cite{ni2025maskgwm} propose MaskGWM, which integrates diffusion-based generation with MAE-style masked reconstruction to significantly improve the long-horizon prediction and generalization ability of driving world models. In the education domain, FMAE~\cite{zhang2024self} proposes a facial masked autoencoder–based pre-training method that learns key facial and spatio-temporal features from unlabeled videos for student engagement recognition. 

In addition, several works explore the synergy between reconstruction and contrastive learning, such as C2MAE~\cite{li2025cross} and CM3AE~\cite{wu2025cm3ae}, which balance instance discrimination and local perception by jointly modeling contrastive objectives and masked image modeling. Other studies focus on pre-training models tailored to specific targets or scenarios, including BYOL~\cite{lu2024cross}, and PLIP~\cite{zuo2023plip} for human-centric perception, S3D~\cite{wald2025revisiting} and HDXrayMAE~\cite{wang2024HDXrayMAE} for medical image perception, and RoMA~\cite{wang2025roma} and RS-vHeat~\cite{hu2025rs} for remote sensing applications.
Different from prior works, our VehicleMAE~\cite{wang2024structural} and its enhanced version VehicleMAE-V2 specifically target vehicle-centric perception by systematically exploiting structural and semantic priors for pre-training.

\subsection{Vehicle-centric Perceptron}  
In recent years, many researchers dedicate to studying vehicle-centric intelligent visual perception problems. Among them, the most widely researched visual tasks include vehicle detection, vehicle attribute recognition, and vehicle re-identification. In the domain of vehicle detection, Zhu et al.~\cite{zhudeformable} introduce a multi-scale deformable attention module, effectively boosting model inference speed. Li et al.~\cite{li2022exploring} propose SFP, which constructs a feature pyramid solely from single-scale feature maps, eliminating hierarchical constraints on the detector backbone network. Wu et al.~\cite{wu2024vfm} propose the VAtt2Vec module to predict and learn unified attribute representations, mitigating differences between visual features and semantic labels.

In the domain of vehicle attribute recognition, Cheng et al.~\cite{cheng2022simple} propose a multimodal framework based on visual-language fusion, aiming to extract inherent textual information from attribute annotations. Wu et al.~\cite{wu2024selective} introduce the Selective Feature Activation Method (SFAM) and the Orthogonal Feature Activation Loss to alleviate the issue of imbalanced attribute distributions in attribute recognition tasks. Bui et al.~\cite{bui2024c2t} propose the C2T-Net attribute recognition network, which effectively integrates SwinT and ViT to better capture both local and global features of individuals. For vehicle re-identification, He et al.~\cite{he2021transreid} introduce the first Transformer-based re-ID framework, which enhances the extraction of global information and fine-grained features. Li et al.~\cite{li2024day} propose a cross-domain class awareness module to mitigate domain discrepancies between day and night images, effectively improving vehicle re-identification performance in day-night alternating traffic scenarios. Shen et al.~\cite{shen2023triplet} present a triplet contrastive representation learning (TCRL) framework that utilizes cluster features to connect part and global features, addressing the issue of unreliable pseudo-labels caused by imbalanced feature information in unsupervised vehicle re-identification. 
Different from these works, in this paper, we propose a vehicle-centric pre-trained large model that supports multiple downstream vehicle-related tasks and achieves higher performances.

\begin{figure*}
\centering
\includegraphics[width=0.95\textwidth]{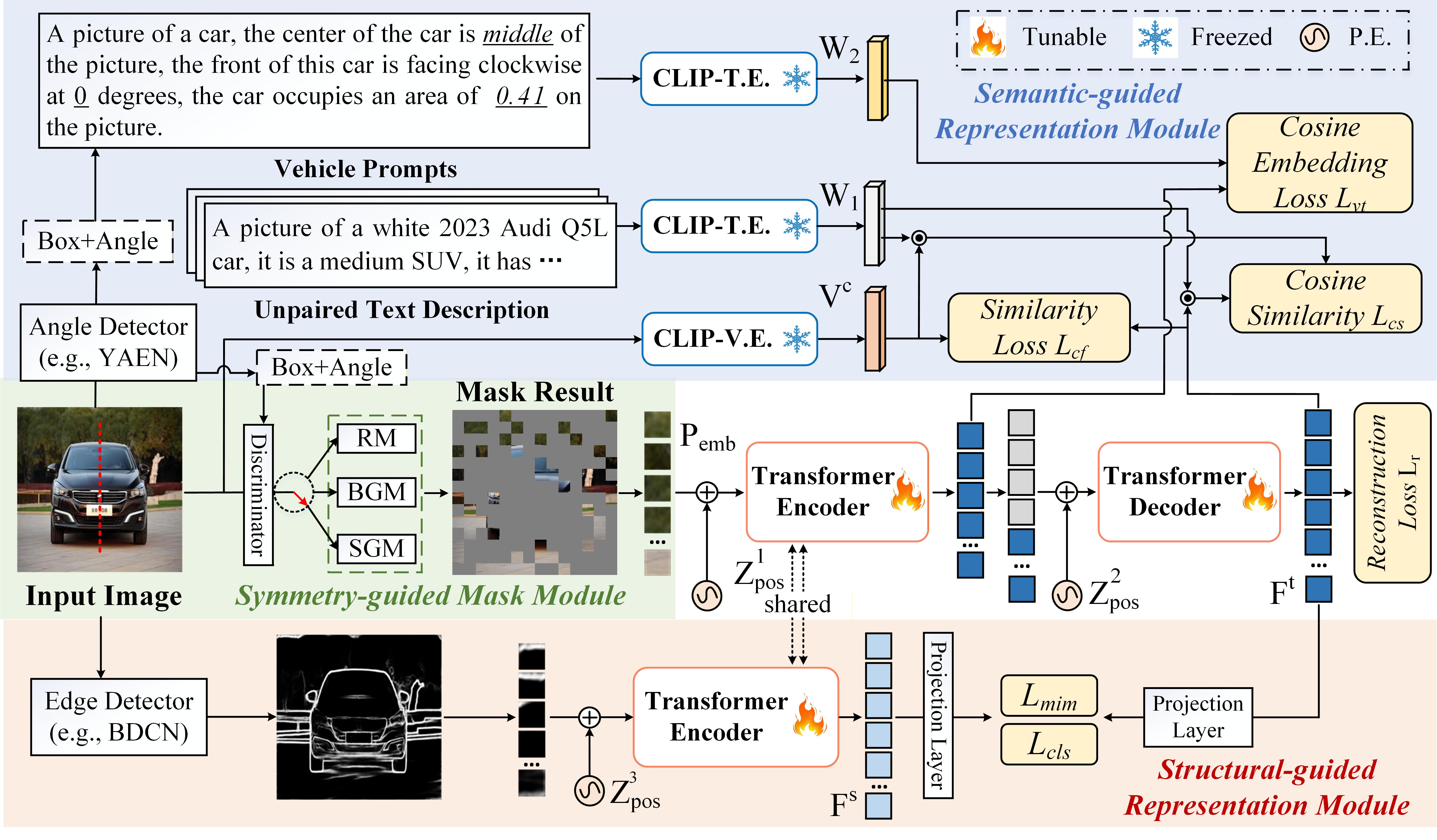}
\caption{An overview of our proposed structured pre-training framework for vehicle perception, termed VehicleMAE-V2. The framework is built upon the structural and semantic information of vehicle targets and consists of three key modules: a Symmetry-guided Mask Module that exploits vehicle symmetry priors to guide the masking strategy; a Contour-guided Representation Module that explicitly models and constrains the spatial structural representation of vehicles; and a Semantics-guided Representation Module that aligns visual features with semantic information to enhance the model’s semantic understanding capability.
} 
\label{fig:framework} 
\end{figure*}

\section{Methodology}  \label{sec3} 

In this section, we first outline our proposed VehicleMAE-V2 framework. Next, we provide a detailed introduction to the network architecture and loss functions, focusing on the Symmetry-guided Mask Module, Contour-guided Representation Module, and Semantics-guided Representation Module. Finally, we discuss five downstream tasks to validate the effectiveness and generalization capability of our proposed VehicleMAE-V2 large model.

\subsection{Overview} 
In this paper, we propose the vehicle-centric pre-training framework based on the Masked Auto-Encoder (MAE), termed VehicleMAE-V2. As shown in Figure~\ref{fig:framework}, given the input vehicle image, we first partition it into non-overlapping patches and mask them with a high ratio by following MAE. 
Different from the widely used random masking strategy, a novel symmetry-guided mask module is proposed in this paper. The underlying insight of this mechanism is that vehicles inherently possess structured information, such as left-right symmetry. The remaining visible patches are projected into tokens and position encoding is added to them before feeding them into the Transformer encoder network. A set of randomly initialized tokens is incorporated by concatenating them with the processed visible tokens. Then, a Transformer decoder is utilized for masked token reconstruction based on the mean squared error loss function. 
However, it is evident that this loss function fails to effectively model the overall structural information of the vehicle, often resulting in blurry reconstructed vehicle images and sub-optimal performance on downstream tasks. In VehicleMAE-V2, considering that vehicle contour lines are predominantly composed of straight lines or curves, we design a contour-guided representation module. By minimizing the probability distribution divergence between contour features and reconstructed features, this module guides the reconstruction process to preserve the overall structural information of vehicles. 

On the other hand, relying solely on the masked reconstruction strategy would lead to a lack of semantic understanding in the model, resulting in part feature confusion issues. Therefore, we design a semantics-guided representation module. First, we introduce a pre-trained CLIP model to encode visual and textual features. We then perform intra-modal and cross-modal knowledge distillation between these CLIP features and the decoder's reconstructed features. This allows our model to learn CLIP's robust representation space and acquire rich vehicle knowledge. Then, we enhance the model's semantic perception of image content by aligning visual features with vehicle-specific textual prompt features.
Extensive experiments on five downstream tasks fully validate the effectiveness of our proposed VehicleMAE-V2. More details are introduced in the following sub-sections.

\subsection{Pre-trained Large Model: VehicleMAE-V2}  

\noindent $\bullet$ \textbf{Symmetry-guided Mask Module.} 
Given the vehicle image $I \in \mathbb{R}^{224 \times 224 \times 3}$, we follow the standard ViT-Base setting and partition the image into non-overlapping patches of size $16 \times 16$. Specifically, the stride is set to 16 along both the height and width dimensions, resulting in $14 \times 14 = 196$ image patches $P_i \in \mathbb{R}^{16 \times 16 \times 3}, i \in \{1, 2, \ldots, 196\}$.
Subsequently, a high ratio of these patches is masked to facilitate effective self-supervised learning by following MAE. Unlike conventional random mask, in this paper, we consider that vehicles are generally laterally symmetrical, which can be used to guide the masking procedure. Specifically, for each input image $I$, we first use the YAEN~\cite{huang2022deep} vehicle angle detector to obtain both location and angle information. Note that there are also some images where the detector is unable to ascertain the vehicle’s orientation, resulting in outputs that are either blank or include only the bounding box information. Consequently, the images are categorized using three types of labels: without any labels, with only a bounding box label, and with both bounding box and angle labels. 

\begin{figure*}
\centering
\includegraphics[width=0.9\textwidth]{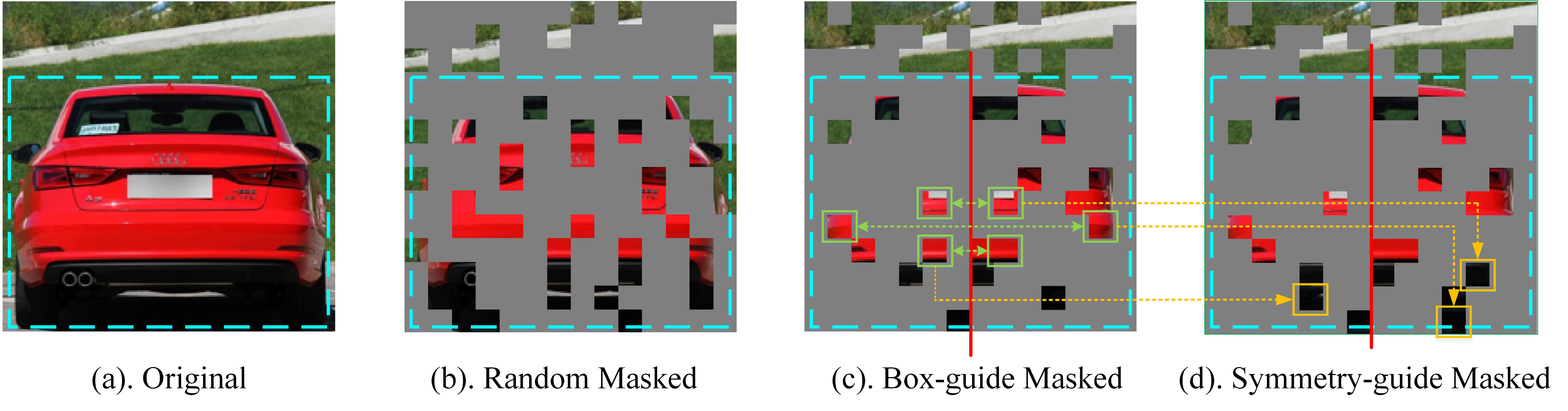}
\caption{Comparison of different masked strategies used in our VehicleMAE-V2 framework, i.e., (b). the random masking, (c). the box-guided masking, and (d). the symmetry-guided masking. 
The blue dotted line represents the box of the vehicle target. The red straight line is the axis of symmetry of the vehicle. The green box and double-headed arrow dashed line indicate the unmasked symmetric patch in the box-guide mask strategy. The yellow box and single-arrow dashed line indicate the result after eliminating symmetrical blocks in the symmetry-guide mask strategy.} 
\label{fig:mask} 
\end{figure*}

In our symmetry-guided mask module, three independent masking strategies are implemented to address different annotation scenarios: the random mask, the box-guided mask, and the symmetry-guided mask, as depicted in Figure~\ref{fig:mask}. Specifically, we adopt the random mask strategy for the unlabeled image by following MAE. We select the box-guided mask strategy for the images with predicted box labels and set different mask ratios for the vehicle and background regions. For input images with both box and angle labels, we use a symmetry-guided mask strategy, which builds on the box-guided mask strategy by adding constraints related to symmetric target blocks. Specifically, we first calculate the symmetric block indices within the target region based on the angle and box and then traverse the masked status of image blocks according to these indices. At least one of the two symmetric blocks is masked. If both symmetric patches are not masked, we randomly mask one patch from them. To maintain the overall masking rate, we randomly remove the mask from one patch selected from a candidate replacement queue. 

After masking $75\%$ of the image patches using the symmetry-guided mask module, the remaining $25\%$ of the patches are fed into the neural network. A convolutional layer with a kernel size $16 \times 16$ is used to map each image patch $P_i$ to a token embedding $P^j_{emb} \in \mathbb{R}^{1 \times 768}, j \in \{1, 2, ..., 49\}$, where 768 denotes the output dimensionality of the convolutional layer and 49 represents the number of unmasked image patches. By concatenating the CLS-token $P_{cls} \in \mathbb{R}^{1 \times 768}$, we obtain the input token embeddings $P_{emb} \in \mathbb{R}^{50 \times 768}$, where 50 consists of one CLS-token and 49 unmasked image tokens. At the same time, we introduce the position encoding $Z^1_{pos} \in \mathbb{R}^{50 \times 768}$ to encode the spatial coordinates of input tokens. Therefore, we have $\tilde{P}_{emb} \in \mathbb{R}^{50 \times 768} = Z^1_{pos} + P_{emb}$ and feed them into the Transformer encoder, which contains 12 Transformer blocks. The output $\bar{P} \in \mathbb{R}^{50 \times 768}$ from the encoder has the same dimensions as the input tokens. After passing the encoder output $\bar{P}$ through a 512-dimensional linear projection layer, it is projected into the embedding $\bar{P}_{emb} \in \mathbb{R}^{50 \times 512}$. 
The randomly initialized masked tokens $P_{mask} \in \mathbb{R}^{147 \times 512}$ and the projected visible tokens $\bar{P}_{emb}$ are first concatenated to form the input $P_{demb} \in \mathbb{R}^{197 \times 512}$ of the decoder network, where 147 denotes the number of masked image patches. A new position encoding $Z^2_{pos} \in \mathbb{R}^{197 \times 512}$ is also incorporated into the input embedding $P_{demb}$. The decoder network contains 8 Transformer blocks, which are used only during the pre-training phase for reconstructing the masked image patches.

The Mean Squared Error (MSE) over pixel space between the masked regions in the original image and the predicted image is adopted as the reconstruction loss to optimize the symmetry-guided mask module, which can be expressed as:
\begin{equation}
L_{r} = \frac {1} {N_{m}} \sum_{t\in P_{m}} \lVert V_{t} - V^r_{t} \lVert_{2},
\end{equation}
where $V$ is the RGB pixel value of the input image, and $V^r$ is the predicted pixel value. $N_{m}$ is the number of masked pixels, $P_{m}$ is the index of the masked pixel, $\lVert * \rVert_2$ refers to the $L_{2}$ loss function.

\noindent $\bullet$ \textbf{Contour-guided Representation Module.} 
In the standard MAE framework, the pixel-wise Mean Squared Error (MSE) loss function is adopted to supervise the reconstruction of masked tokens. Obviously, this loss function can achieve good results in guiding fine-grained pixel-level prediction, but it falls short in modeling the structure information of the overall vehicle. The structure prior information of vehicles is significant, including lines, curves, etc. In addition to the significant shape information of different components, there are also strict positional relationships between the components. Therefore, we consider utilizing such information to achieve higher-quality masked token reconstruction.

As shown in Figure~\ref{fig:framework}, we first utilize the edge detector BDCN~\cite{he2020bdcn} to get the contour map of the input vehicle image. Then, we partition it into non-overlapping patches and project them into tokens $P_{lk} \in \mathbb{R}^{197 \times 768}$. The Transformer network used for the encoding of raw vehicle image is adopted to encode the contour map. The output feature vectors serve as the supervision for vehicle reconstruction. 
Specifically, we project the skeleton tokens and reconstructed invisible tokens into probability distributions with K dimensions (i.e., $P^{patch}_{\theta^{\prime}}(F^s_{i})$ and $P^{patch}_{\theta} (F^t_{i})$ in Eq.~\ref{LmimLoss}, respectively) using two separate projection layers, 

\begin{equation}
\label{LmimLoss}
L_{mim} = -\sum^{N_p}_{i=1} P^{patch}_{\theta^{\prime}}(F^s_{i})^Tlog P^{patch}_{\theta} (F^t_{i}), 
\end{equation} 
where $N_p$ is the number of masked patches in the encoding phase, $\theta^{\prime}$ and $\theta$ are the parameters of the two separate projection layers, $F^{s}$ is the skeleton feature output by the contour guided representation module, and $F^t$ is the feature corresponding to the reconstructed invisible token. 

In order to obtain better visual semantic information, we also project the skeleton CLS token and reconstructed CLS token to derive their respective classification distributions. The similarity loss function can be formally written as: 
\begin{equation}
\label{LclsLoss}
L_{cls} =  P^{cls}_{\theta^{\prime}}(F^s_{cls})^Tlog P^{cls}_{\theta} (F^t_{cls}).
\end{equation}
where $F^s_{cls}$ and $F^t_{cls}$ are the class-tokens output by the contour guided representation module and the reconstruction decoder, respectively.

\begin{table*}[!htp]
\centering 
\caption{Experimental results of our model, other pre-trained models, and task-specific models on vehicle attribute recognition (VAR), vehicle detection (V-Det), vehicle re-identification (V-Reid), vehicle fine-grained recognition (VFR), and vehicle part segmentation (VPS).} 
\label{table1} 
\footnotesize
\resizebox{1\textwidth}{!}{
\begin{tabular}{cc|ccccc|ccc|cc|c|cc}  
\hline
\multirow{2}{*}{\raggedright \textbf{Method}} 
&\multirow{2}{*}{\raggedright \textbf{Dataset}} 
&\multicolumn{5}{c|}{\textbf{VAR}} 
&\multicolumn{3}{c|}{\textbf{V-Det}}
&\multicolumn{2}{c|}{\textbf{V-Reid}}
&\multicolumn{1}{c|}{\textbf{VFR}}
&\multicolumn{2}{c}{\textbf{VPS}} \\
& &$mA$ &$Acc$ & $Prec$ & $Recall$ & $F_1$ & $AP_{[0.5:0.95]}$  & $AP_{0.5}$ &  $AP_{0.75}$ & $mAP$ & $R1$ & $Acc$ & $mIou$ & $mAcc$   \\ 
\hline 
PromptPAR~\cite{wang2024pedestrian}   &-     &90.58 &94.10 &95.77 &95.78 &95.58    &- &- &-   &- &-  &-   &- &- \\
VFM-Det~\cite{wu2024vfm}   &-     &- &- &- &- &-    &46.9 &66.5 &51.6   &- &-  &-   &- &- \\
KMINet~\cite{liu2025knowledge}   &-     &- &- &- &- &-    &- &- &-   &85.7 &97.7   &-   &- &- \\
DIFFUSEMIX~\cite{islam2024diffusemix}   &-     &- &- &- &- &-    &- &- &-   &- &-   &93.1   &- &- \\
GCNet~\cite{yang2025golden}   &-     &- &- &- &- &-    &- &- &-   &- &-   &-   &73.82 &79.31 \\
\hline 
Scratch   &-     &84.67 &80.86 &84.66 &85.77 &84.90    &41.2 &59.6 &44.5   &35.3 &57.3   &24.8   &49.36 &59.22 \\
MoCov3~\cite{MoCov32021},  &ImagNet-1K    &90.38 &93.88 &95.57 &95.48 &95.33   &45.0 &65.6 &49.3   &75.5 &94.4   &91.3  &73.17 &78.60 \\
DINO~\cite{caron2021emerging}  &ImagNet-1K     &89.92 &91.09 &92.84 &93.60 &93.11   &44.1 &64.8 &48.3   &64.3 &91.5   & -  &68.43 &73.37  \\
IBOT~\cite{zhouimage}  &ImagNet-1K     &89.51 &90.17 &91.95 &93.03 &92.37   &41.8 &60.8 &46.5   &68.9 &92.6   &81.1  &66.03 &71.06  \\
\hline 
MAE~\cite{he2022mae}  &ImagNet-1K     &89.69 &93.60 &94.81 &95.54 &95.08   &42.4 &61.8 &46.4   &76.7 &95.8   &91.2  &69.54 &75.36 \\
MAE~\cite{he2022mae}  &Autobot1M     &90.19 &94.06 &95.45 &95.68 &95.43   &43.7 &63.0 &47.8   &75.5 &95.4   &91.3  &69.00 &75.36 \\
VehicleMAE~\cite{wang2024structural}  &Autobot1M     &92.21 &94.91 &96.00 &96.50 &96.17   &46.9 &66.5 &51.6   &85.6 &97.9   &94.5   &73.29 &80.22  \\
VehicleMAE-V2  &Autobot1M    &92.43 &95.67 &96.56 &97.00 &96.70   &47.3 &67.2 &52.9   &86.6 &98.0   &94.8   &74.66 &80.67 \\
VehicleMAE-V2  &Autobot4M    &\textbf{93.19} &\textbf{95.91} &\textbf{96.89} &\textbf{97.28} &\textbf{96.98}   &\textbf{48.5} &\textbf{67.4} &\textbf{53.8}   &\textbf{87.3} &\textbf{98.0}   &\textbf{94.9}   &\textbf{75.04} &\textbf{81.04}  \\
\hline 
\end{tabular}} 
\end{table*}

\noindent $\bullet$ \textbf{Semantics-guided Representation Module.} 
The MAE framework is trained through original image reconstruction, and our newly proposed symmetry-guided mask module helps select the most valuable image patches to be masked. The contour-guided representation module assists the model in better reconstructing the spatial layout of the vehicle. Although this achieves good performance, the model still lacks understanding of the image content and vehicle targets. Therefore, we introduce two types of text modal information to enrich the model with semantic information, enabling the large model to better understand the vehicles. 

In this work, we collect attribute information for different vehicles to construct a set of textual descriptions that are not completely paired with images. We use the pre-trained vision-language model CLIP~\cite{radford2021learning} to process the vehicle image $I$ and the textual description $T^{ua} = [w_1, w_2, ..., w_m]$, where $w_i$ represents the $i^{th}$ textual description. Specifically, the CLIP~\cite{radford2021learning} language encoder embeds each text $w_i$ to obtain the tokens $\widetilde{w_i} \in \mathbb{R}^{1 \times 512}$. The CLIP text feature $W =  [\widetilde{w_1}, \widetilde{w_2}, ..., \widetilde{w_m}]$. Meanwhile, the CLIP visual encoder transforms the vehicle image into visual tokens $V^c \in \mathbb{R}^{1 \times 512}$. Note that both the CLIP visual and language encoders have fixed parameters. To leverage the feature extraction capabilities of the pre-trained CLIP visual encoder, we introduce a similarity loss between the CLIP visual features and the MAE Transformer decoded features. The formula is as follows:
\begin{equation}
L_{cf} = (\frac {F^t} {\lVert F^t \lVert_{2}}- \frac {V^c} {\lVert V^c \lVert_{2}})^2  = (\widetilde{F^t} - \widetilde{V^c} )^2,
\end{equation}
where $F^t$ is decoded features from MAE, and $\lVert\mathbf{*}\lVert_2$ refers to the $L_{2}$ norm.

Then, we also consider the consistency constraint between the similarity distribution of CLIP text-visual features and CLIP text-MAE decoded visual features. To be specific, the similarity between the CLIP text $\widetilde{w_{j}}, j=\{1, 2, ..., m\}$ and the visual features is firstly calculated by: 
\begin{equation}
s_{j}(\widetilde{V^c},\widetilde{w_{j}}) = \frac {exp((\widetilde{V^c}*\widetilde{w_{j}})/ \tau)} {\sum^m_{n=1} exp((\widetilde{V^c}*\widetilde{w_{n}})/ \tau)},
\end{equation} 
where $\tau$ is a temperature hyper-parameter, which is set to 1 in our experiments. 
The similarity distribution between the text and CLIP visual features can be obtained via:
\begin{equation}
S(\widetilde{V^c},W) = [s_{1}(\widetilde{V^c},\widetilde{w_{1}}),...,s_{m}(\widetilde{V^c},\widetilde{w_{m}})].
\end{equation}
Similarly, we can get the similarity distribution between the text and MAE decoded visual features via: 
\begin{equation}
S(\widetilde{F^t},W) = [s_{1}(\widetilde{F^t},\widetilde{w_{1}}),...,s_{m}(\widetilde{F^t},\widetilde{w_{m}})]. 
\end{equation}
We realize the consistency constraint by minimizing the cross-entropy loss between the two similarity distributions.

\begin{figure*}[!htp]
\center
\includegraphics[width=\textwidth]{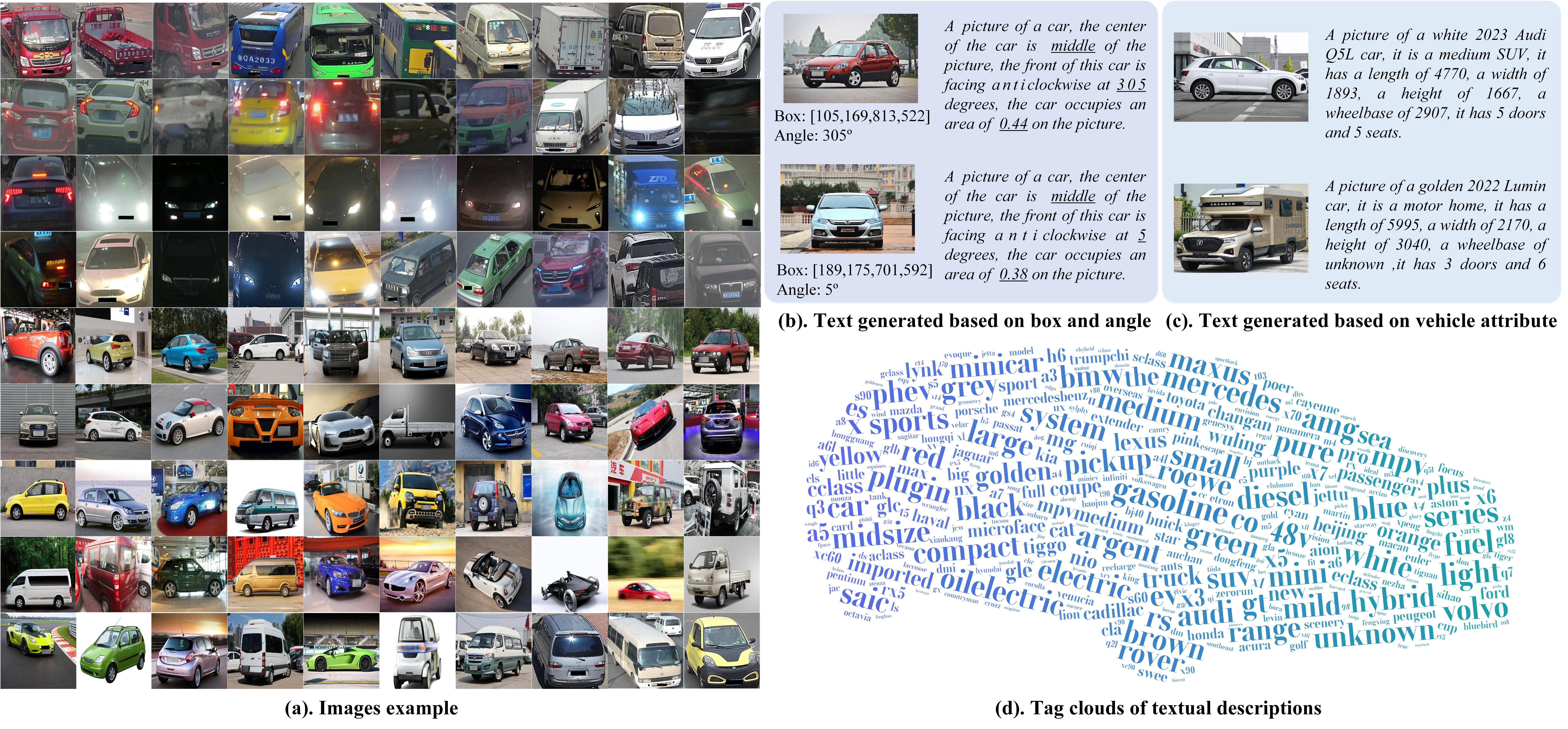}
\caption{Representative samples of our proposed Autobot4M dataset.} 
\label{fig:dataSamples}
\end{figure*}

To enhance our model, we introduce a regularization term for the similarity distribution between the MAE decoded features and CLIP text features. Therefore, the regularized similarity distribution consistency loss function can be formulated as: 
\begin{equation}
\begin{split}
L_{cs} &= KL(S(\widetilde{V^c},W),S(\widetilde{F^t},W))  +  H(S(\widetilde{F^t},W)), 
\end{split}
\end{equation}
where $H$ denotes the entropy, and $KL$ is the kullback-leibler divergence. 

On the other hand, the YAEN~\cite{huang2022deep} vehicle angle detector can detect box and angle labels for most images. Thus, we utilize these two labels to generate textual descriptions that correspond to each image. Specifically, there are three scenarios for each input image based on the output of the YAEN~\cite{huang2022deep}, i.e., no label, with box label, and with both box and angle labels. 

For images without predicted labels, we don't generate textual descriptions for these images used in contrastive learning. For images with box labels, we calculate the center position and the proportion of the vehicle object in the image based on the box information. The center position $c$ is the center point of the box, and the proportion $r$ is the ratio of the box area to the image area. The large language model ChatGPT is utilized to generate a template $Q_{1}$, and the center position and proportion information are embedded into the template to generate a content-relevant textual description $w^{a}_{i}=Q_{1}(c_{i},r_{i})$ for that image. 

For images with both box and angle labels, we introduce the center position, proportion, and vehicle angle $a$ and generate a template $Q_{2}$ using the ChatGPT model.  
Then, a textual description $w^{a}_{j}=Q_{2}(c_{j},r_{j},a_{j})$ can be generated based on the template. Thus, we can get the content-relevant textual descriptions $T^{a}=[w^{a}_{1},w^{a}_{2},...,w^{a}_{u}]$ for most images. As shown in Figure~\ref{fig:dataSamples} (b), we provide examples of the corresponding generated textual descriptions.
In this framework, we employ a pre-trained CLIP language encoder to process these content-relevant textual descriptions. Specifically, the textual description $w^{a}_{i}$ is input into the CLIP text encoder to obtain text features $F^{w}_i \in \mathbb{R}^{1 \times 512}$. It is important to note that the parameters of both the CLIP visual and language encoders are fixed. Then we normalize the image features $\bar{P}$ output by the encoder in the symmetry-guided mask module and the text features $F^{w}$ using the L2 norm and calculate the cosine embedding loss between the normalized features. The formula is as follows:
\begin{equation}
L_{vt} = \frac {1} {N_b} \sum^{N_b}_{i=1} CEL(\frac {\bar{P}_i} {\lVert \bar{P}_i \lVert_{2}}, \frac {F^w_i} {\lVert F^w_i \lVert_{2}}). 
\end{equation}
where $CEL$ stands for Cosine Embedding Loss, $N_b$ represents the batch size, and $\lVert \mathbf{*} \lVert_2$ refers to the $L_{2}$ norm.

Finally, the overall loss function of our proposed VehicleMAE-V2 can be expressed as:
\begin{equation}
L = \lambda_{r}L_{r} + \lambda_{mim}L_{mim} +\lambda_{cls}L_{cls} + \lambda_{cf}L_{cf} + \lambda_{cs}L_{cs} + \lambda_{vt}L_{vt},
\end{equation}
where $\lambda_{r}, \lambda_{mim}, \lambda_{cls}, \lambda_{cf}, \lambda_{cs},$ and $\lambda_{vt}$ are hyperparameters controlling the relative importance of each loss component.

\subsection{Downstream Tasks} 

In this paper, we validate the effectiveness and generalization capability of our proposed large model VehicleMAE-V2 across five vehicle perception tasks. These tasks include object detection, part segmentation, attribute recognition, re-identification, and fine-grained recognition. A brief introduction to these tasks can be found in our supplementary materials.


\begin{table*}[!htbp] 
\centering 
\caption{Ablation study on Symmetry-guided Mask Module (SMM) and loss functions in Contour-guided Representation Module (CRM) and Semantics-guided Representation Module (SRM).}  
\label{AblationStudy1}  
\footnotesize
\resizebox{1\textwidth}{!}{
\begin{tabular}{c|c|cc|ccc|c|ccccccc} 
\hline 
\multirow{2}{*}{\raggedright \textbf{M.R.}}
&\raggedright \textbf{MAE Loss}
&\multicolumn{2}{c|}{\textbf{CRM}} 
&\multicolumn{3}{c|}{\textbf{SRM}}
&\multirow{2}{*}{\raggedright \textbf{SMM}} 
&\multicolumn{5}{c}{\textbf{VAR}} &\multicolumn{2}{c}{\textbf{VPS}} \\
&$L_{r}$ &$L_{mim}$ &$L_{cls}$ &$L_{cs}$ &$L_{cf}$ &$L_{vt}$ & & $mA$ & $Acc$ & $Prec$ & $Recall$ & $F_1$ & $mIou$ & $mAcc$ \\ 
\hline 
\multirow{9}{*}{\raggedright $75\%$}
&\checkmark & & & & & & &90.19 &94.06 &95.45 &95.68 &95.43   &69.00 &75.36 \\
&\checkmark &\checkmark & & & & & &91.27 &94.11 &95.29 &95.82 &95.50   &70.34 &75.70 \\
&\checkmark &\checkmark &\checkmark & & & & &91.71 &94.54 &95.65 &96.28 &95.88   &70.65 &76.04 \\
&\checkmark & & &\checkmark & & & &92.12 &94.28 &95.42 &96.23 &95.71   &71.90 &76.47 \\
&\checkmark & & &\checkmark &\checkmark & & &92.15 &94.58 &95.69 &96.36 &95.92   &71.87 &77.93 \\
&\checkmark &\checkmark &\checkmark &\checkmark &\checkmark & & &92.21 &94.91 &96.00 &96.50 &96.17   &73.29 &80.22 \\
&\checkmark &\checkmark &\checkmark &\checkmark &\checkmark& \checkmark& &92.66 &95.23 &96.26 &96.72 &96.42   &74.32 &80.41 \\
&\checkmark &\checkmark &\checkmark &\checkmark &\checkmark& &\checkmark &92.37 &95.46 &96.40 &96.89 &96.58   &74.13 &80.21 \\
&\checkmark &\checkmark &\checkmark &\checkmark &\checkmark &\checkmark &\checkmark &92.43 &\textbf{95.67} &\textbf{96.56} &\textbf{97.00} &\textbf{96.70}   &\textbf{74.66} &\textbf{80.67} \\
\hline
\multirow{3}{*}{\raggedright $85\%$}
&\checkmark &\checkmark &\checkmark &\checkmark &\checkmark & & &90.73 &94.18 &95.32 &95.97 &95.55   &70.91 &77.12 \\
&\checkmark &\checkmark &\checkmark &\checkmark &\checkmark & &\checkmark &91.50 &94.55 &95.76 &96.20 &95.88   &71.24 &77.67 \\
&\checkmark &\checkmark &\checkmark &\checkmark &\checkmark&\checkmark& \checkmark &\textbf{93.01} &95.47 &96.40 &96.97 &96.59   &71.86 &78.14 \\
\hline
\multirow{2}{*}{\raggedright \textbf{}} 
&\multirow{2}{*}{\raggedright \textbf{}} 
&\multirow{2}{*}{\raggedright \textbf{}} 
&\multirow{2}{*}{\raggedright \textbf{}} 
&\multirow{2}{*}{\raggedright \textbf{}} 
&\multirow{2}{*}{\raggedright \textbf{}} 
&\multirow{2}{*}{\raggedright \textbf{}} 
&\multirow{2}{*}{\raggedright \textbf{}} 
&\multicolumn{2}{c}{\textbf{V-ReID}} &\textbf{VFR} &\multicolumn{3}{c}{\textbf{V-Det}}\\
& & & & & & & & $mAP$ & $R1$ & $Acc$ &\textbf{$AP_{[0.5:0.95]}$} &\textbf{$AP_{0.5}$} & \textbf{$AP_{0.75}$} \\ 
\hline
\multirow{9}{*}{\raggedright $75\%$}
&\checkmark & & & & & & &75.5 &95.4   &91.3  &43.7 &63.0 &47.8   & \\
&\checkmark &\checkmark & & & & & &79.7 &96.1   &93.2   &44.3 &62.7 &49.5   & \\
&\checkmark &\checkmark &\checkmark & & & & &83.4 &96.6   &93.7   &45.0 &63.3 &50.6   & \\
&\checkmark & & &\checkmark & & & &84.1 &97.1   &94.1   &45.6 &64.7 &49.9 & \\
&\checkmark & & &\checkmark &\checkmark & & &85.2 &97.1   &94.3   &46.3 &65.3 &51.6   & \\
&\checkmark &\checkmark &\checkmark &\checkmark &\checkmark & & &85.6 &97.9   &94.5   &46.9 &66.5 &51.6 \\
&\checkmark &\checkmark &\checkmark &\checkmark &\checkmark &\checkmark & &86.1 &97.9   &94.7   &47.1 &66.9 &52.2 \\
&\checkmark &\checkmark &\checkmark &\checkmark &\checkmark & &\checkmark &86.0 &97.9   &94.7   &47.2 &66.3 &52.0 \\
&\checkmark &\checkmark &\checkmark &\checkmark &\checkmark &\checkmark&\checkmark  &\textbf{86.6} &\textbf{98.0}   &\textbf{94.8}   &\textbf{47.3} &\textbf{67.2} &\textbf{52.9} \\
\hline
\multirow{3}{*}{\raggedright $85\%$}
&\checkmark &\checkmark &\checkmark &\checkmark &\checkmark & & &82.1 &96.3   &93.5   &44.9 &64.1 &50.5 \\
&\checkmark &\checkmark &\checkmark &\checkmark &\checkmark & &\checkmark &83.1 &97.1   &93.8   &45.2 &63.4 &50.8 \\
&\checkmark &\checkmark &\checkmark &\checkmark &\checkmark & \checkmark&\checkmark  &83.6 &97.3   &94.3   &46.3 &64.6 &52.3 \\
\hline
\end{tabular}}
\end{table*}

\begin{table*}[!htbp] 
\centering 
\caption{Ablation Study on Loss Weight Settings.}  
\label{AblationStudy5}  
\footnotesize
\resizebox{1\textwidth}{!}{
\begin{tabular}{c|ccccc|ccc|cc|c|cc} 
\hline 
\multicolumn{1}{c|}{\textbf{Loss Weight Settings}} 
&\multicolumn{5}{c|}{\textbf{VAR}} 
&\multicolumn{3}{c|}{\textbf{V-Det}}
&\multicolumn{2}{c|}{\textbf{V-ReID}} 
&\textbf{VFR}
&\multicolumn{2}{c}{\textbf{VPS}} \\
$L_{r},L_{mim},L_{cls},L_{cs},L_{cf},L_{vt}$&$mA$ &$Acc$ & $Prec$ & $Recall$ & $F_1$ 
& $AP_{[0.5:0.95]}$  & $AP_{0.5}$ &  $AP_{0.75}$ 
& $mAP$ & $R1$ 
& $Acc$ 
& $mIou$ & $mAcc$   \\ 
\hline

(1,1,1,1,1,1)&    91.67& 94.63& 95.85& 96.30& 95.97&   44.5& 63.7& 49.3&   82.4& 97.1&   93.7&   70.73& 77.01 \\
(1,0.02,0.02,1,1,1)&   91.82& 95.32& 96.52& 96.58& 96.45&   46.4& 65.9& 51.7&   85.9& 97.9&   94.7&   73.54& 79.32 \\
(4,0.02,0.02,1,1,1)&    \textbf{92.43} &\textbf{95.67} &\textbf{96.56} &\textbf{97.00} &\textbf{96.70}    &\textbf{47.3} &\textbf{67.2} &\textbf{52.9}   &\textbf{86.6} &\textbf{98.0}   &\textbf{94.8}    &\textbf{74.66} &\textbf{80.67} \\
\hline
\end{tabular}}
\end{table*}

\section{Pre-training Dataset: Autobot4M} \label{sec4}

In this work, we collect a large-scale pre-training dataset for vehicle-centric perception, called Autobot4M, to support the training of our proposed VehicleMAE-V2. The dataset contains 4,020,368 vehicle images from various scenes and 12,693 textual descriptions of different vehicle models. 

The images in this dataset are sourced from our constructed vehicle pre-training dataset Autobot1M, as well as two publicly available vehicle datasets: the PKU-VD~\cite{yan2017exploiting} dataset and the SODA10M~\cite{han2soda10m} dataset. Specifically, Autobot1M includes 1,026,394 vehicle images from different scenes, including 732,112 surveillance images and 294,282 network images. The training set of the PKU-VD dataset contains 1,886,682 individual vehicle images from surveillance scenes, all of which are added to our dataset. The SODA10M dataset consists of 10 million road scene images, from which we select a subset and use the YOLOv5~\footnote{\url{https://github.com/ultralytics/yolov5}} detection algorithm to crop out 1,107,292 individual vehicle images. As shown in Figure~\ref{fig:dataSamples} (a), the images in our dataset fully consider various challenging factors in vehicle perception tasks, such as low light, motion blur, occlusion, and multi-view angles. It also covers a range of capture tools, including surveillance cameras, in-vehicle cameras, handheld cameras, and web-rendered images.

At the same time as crawling vehicle images from the Internet, we also collect eleven attribute information corresponding to the vehicle models, including brand, color, energy type, level, length, width, height, doors, seats, wheelbase, and years. Based on these attributes, we generate corresponding textual descriptions for each vehicle model. 
As shown in Figure~\ref{fig:dataSamples} (c) and Figure~\ref{fig:dataSamples} (d), we provide examples of the textual descriptions and the frequency of word occurrences in these textual descriptions.

\section{Experiments}  \label{sec5}

\subsection{Datasets and Evaluation Metric} \label{DSdatasets}

In our five different downstream tasks, four datasets are adopted for the downstream validation, including the \textbf{VeRi} dataset~\cite{liu2016Veri776}, \textbf{Stanford Cars} dataset~\cite{krause2013Stanfordcars}, and \textbf{PartImageNet} dataset~\cite{he2022partimagenet}, \textbf{Cityscapes} dataset~\cite{cordts2016cityscapes}.
Multiple evaluation metrics are used for different downstream tasks, including $mA$, $Acc$, $Prec$, $Recall$, $F_1$, $mAP$, $R1$, $mIoU$, $mAcc$, $AP_{[0.5:0.95]}$, $AP_{0.5}$, and $AP_{0.75}$. More details can be found in our supplementary materials.

\subsection{Implementation Details} \label{impleDetails} 
In our pre-training phase, the learning rate is set as 0.0002, and the weight decay is 0.04. The AdamW~\cite{adamW} is selected as the optimizer to train our model. The batch size is 512 and trains for a total of 100 epochs on our Autobot1M or Autobot4M dataset. Additional training details are provided in the supplementary materials.

\subsection{Comparison with State-of-the-art Algorithms}   \label{compSOTA}
In this experiment, we systematically validate the effectiveness of the VehicleMAE-V2 large-scale pre-trained model across five downstream tasks, and further analyze the impact of fine-tuning with different amounts of training data on model performance. For each downstream task, we conduct comprehensive comparisons between our proposed method and baseline models, task-specific models, as well as state-of-the-art pre-trained models, with the results summarized in Table~\ref{table1}. Specifically, the compared approaches include models trained from scratch without pre-training; MoCov3~\cite{MoCov32021}, DINO~\cite{caron2021emerging}, IBOT~\cite{zhouimage}, and MAE~\cite{he2022mae} pre-trained on the ImageNet-1K dataset; MAE and VehicleMAE pre-trained on the Autobot1M dataset; and VehicleMAE-V2 pre-trained on the Autobot1M and Autobot4M datasets. In addition, we include several task-specific vehicle models for comparison, including the vehicle attribute recognition model PromptPAR~\cite{wang2024pedestrian}, the vehicle detection model VFM-Det~\cite{wu2024vfm}, the vehicle re-identification model KMINet~\cite{liu2025knowledge}, the fine-grained vehicle recognition model DIFFUSEMIX~\cite{islam2024diffusemix}, and the vehicle part segmentation model GCNet~\cite{yang2025golden}.

\begin{figure*}
\centering
\includegraphics[width=\textwidth]{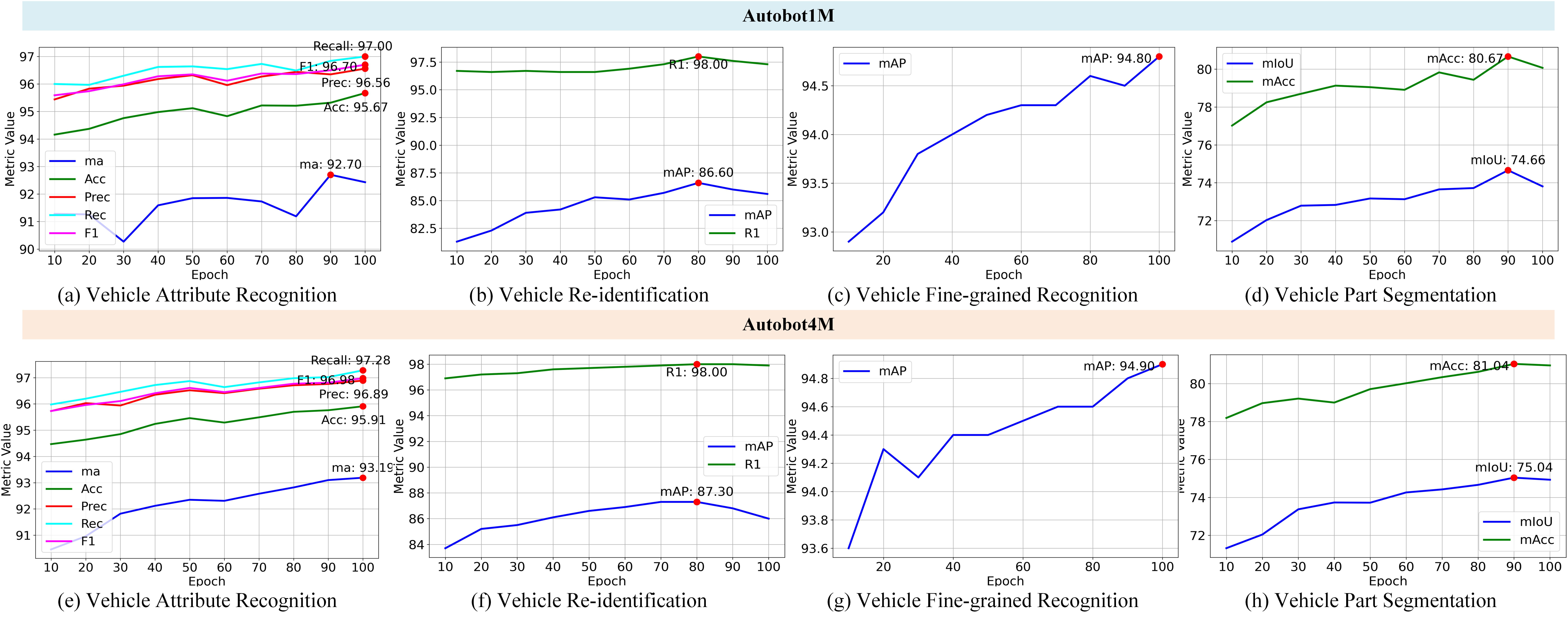}
\caption{The impact of pre-training epochs on experimental results across different downstream tasks.}
\label{fig:epoch} 
\end{figure*}

\begin{figure}
\centering
\includegraphics[width=3.3in]{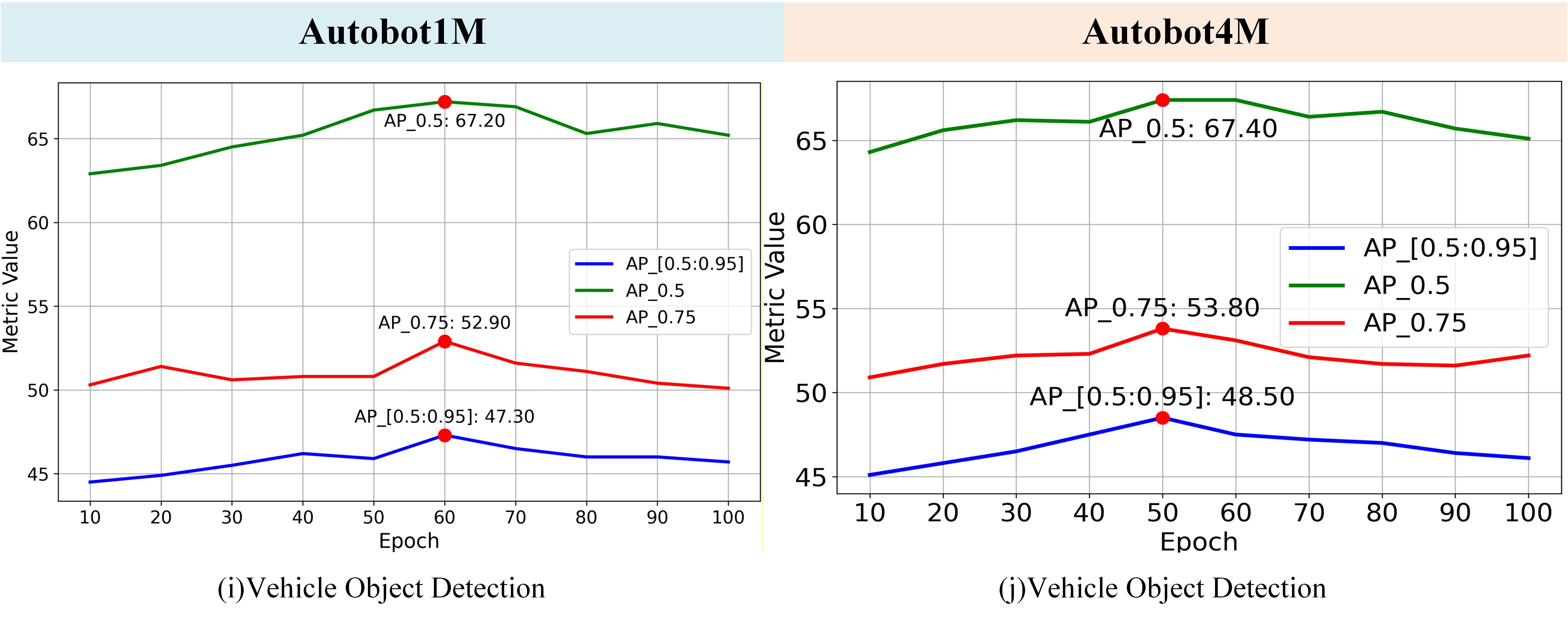}
\caption{The impact of pre-training epochs on experimental results across detection tasks.}
\label{fig:epoch_det} 
\end{figure}

\begin{table*}[!htbp] 
\centering 
\caption{Ablation study on the ratio of masked tokens.}  
\label{AblationStudy4}  
\footnotesize
\resizebox{1\textwidth}{!}{
\begin{tabular}{c|c|ccccc|ccc|cc|c|cc} 
\hline 
\multirow{2}{*}{\raggedright \textbf{Method}} 
&\multirow{2}{*}{\raggedright \textbf{M.R.}} 
&\multicolumn{5}{c|}{\textbf{VAR}}
&\multicolumn{3}{c|}{\textbf{V-Det}}
&\multicolumn{2}{c|}{\textbf{V-ReID}}
&\textbf{VFR}
&\multicolumn{2}{c}{\textbf{VPS}} \\
& &$mA$ &$Acc$ & $Prec$ & $Recall$ & $F_1$ 
& $AP_{[0.5:0.95]}$  & $AP_{0.5}$ &  $AP_{0.75}$ 
& $mAP$ & $R1$ 
& $Acc$ 
& $mIou$ & $mAcc$   \\ 
\hline 
\multirow{4}{*}{\raggedright VehicleMAE} 
&0.25 &90.48 &94.34 &95.63 &96.02 &95.72   &46.3 &66.0 &51.4    &84.9 &97.0   &94.0   &72.31 &78.20 \\
&0.50 &91.88 &94.35 &95.55 &96.11 &95.72   &46.9 &66.8 &52.1   
&85.2 &97.3   &94.3   &71.90 &77.66 \\
&0.75 &92.21 &94.91 &96.00 &96.50 &96.17   &46.9 &66.5 &51.6    &85.6 &97.9   &94.5   &73.29 &80.22 \\
&0.85 &90.73 &94.18 &95.32 &95.97 &95.55   &44.9 &64.1 &50.5    &82.1 &96.3   &93.5   &70.91 &77.12 \\
\hline
\multirow{2}{*}{\raggedright VehicleMAE-V2} 
&0.75 &92.43 &\textbf{95.67} &\textbf{96.56} &\textbf{97.00} &\textbf{96.70}   &\textbf{47.3} &\textbf{67.2} &\textbf{52.9}   &\textbf{86.6} &\textbf{98.0}   &\textbf{94.8}   &\textbf{74.66} &\textbf{80.67} \\
&0.85 &\textbf{93.01} &95.47 &96.40 &96.97 &96.59   &46.3 &64.6 &52.3   &83.6 &97.2   &94.3   &71.86 &78.14 \\
\hline
\end{tabular}}
\end{table*}

\begin{table*}[!htbp] 
\centering 
\caption{Comparative experiments with different backbone networks.}  
\label{AblationStudy5}  
\resizebox{1\textwidth}{!}{
\footnotesize
\begin{tabular}{c|ccccc|ccc|cc|c|cc} 
\hline 

\multirow{2}{*}{\raggedright \textbf{Backbone}} 
&\multicolumn{5}{c|}{\textbf{VAR}} 
&\multicolumn{3}{c|}{\textbf{V-Det}}
&\multicolumn{2}{c|}{\textbf{V-ReID}} 
&\textbf{VFR}
&\multicolumn{2}{c}{\textbf{VPS}} \\

&$mA$ &$Acc$ & $Prec$ & $Recall$ & $F_1$ 
& $AP_{[0.5:0.95]}$  & $AP_{0.5}$ &  $AP_{0.75}$ 
& $mAP$ & $R1$ 
& $Acc$ 
& $mIou$ & $mAcc$   \\ 
\hline 

ViT-B/16 &92.43 &95.67 &96.56 &97.00 &96.70   &47.3 &67.2 &52.9   &86.6 &98.0   &94.8   &74.66 &80.67 \\
ViT-L/16 &\textbf{93.45} &\textbf{95.91} &\textbf{96.99} &\textbf{97.23} &\textbf{96.98}   &\textbf{47.6} &\textbf{67.5} &\textbf{53.2}   &\textbf{89.5} &\textbf{98.3}   &\textbf{95.1}   &\textbf{75.11} &\textbf{81.06} \\
\hline

\end{tabular} }
\end{table*}

\noindent $\bullet$ \textbf{Vehicle Attribute Recognition.}
For the vehicle attribute recognition task, we incorporate different pre-trained models into the baseline VTB~\cite{cheng2022VTB} framework to evaluate the effectiveness of the proposed approach. As shown in Table~\ref{table1}, when initializing VTB with the VehicleMAE model pre-trained on the Autobot1M dataset, the model achieves $92.21\%$, $94.91\%$, $96.00\%$, $96.50\%$, and $96.17\%$ on the $mA$, $Acc$, $Prec$, $Recall$, and $F_1$ metrics, respectively. Replacing the pre-training strategy with the proposed VehicleMAE-V2 consistently improves all evaluation metrics to $92.43\%$, $95.67\%$, $96.56\%$, $97.00\%$, and $96.70\%$. When further scaling the pre-training data to Autobot4M, the model achieves the best performance, obtaining $93.19\%$, $95.91\%$, $96.89\%$, $97.28\%$, and $96.98\%$ on the five metrics, significantly outperforming other general-purpose pre-trained models as well as the task-specific PromptPAR~\cite{wang2024pedestrian} model. It is worth noting that the performance gaps among different methods on this task are relatively small, which may be attributed to the relatively low difficulty of the dataset, where overall performance is close to saturation.

\noindent $\bullet$ \textbf{Vehicle Detection.}
For the vehicle detection task, we fine-tune different pre-trained models within the VFM-Det~\cite{wu2024vfm} framework to assess the effectiveness of the proposed method in object localization and recognition. As reported in Table~\ref{table1}, the VehicleMAE model pre-trained on Autobot1M achieves $46.9\%$, $66.5\%$, and $51.6\%$ on $AP_{0.5:0.95}$, $AP_{0.5}$, and $AP_{0.75}$, respectively. By adopting the VehicleMAE-V2 pre-training strategy, detection performance is further improved, with $AP_{0.5:0.95}$ increasing to $47.3\%$ and $AP_{0.75}$ to $52.9\%$. When the pre-training data scale is expanded to Autobot4M, the model attains the best detection performance, achieving $47.5\%$, $67.4\%$, and $53.1\%$ on the three metrics. These results clearly surpass those of general-purpose pre-trained models and other vehicle detection methods, demonstrating that VehicleMAE-V2 provides more robust representations for detection tasks.

\noindent $\bullet$ \textbf{Vehicle Re-identification.}
For the vehicle re-identification task, we conduct evaluations using the TransREID~\cite{he2021transreid} framework, with a focus on the impact of pre-training models on fine-grained discriminative feature learning. As shown in Table~\ref{table1}, when employing the general-purpose MAE pre-training on the Autobot1M dataset, the model achieves $75.5\%$ mAP and $95.4\%$ Rank-1 (R1) accuracy. In contrast, the proposed VehicleMAE-V2 yields substantial improvements under the same pre-training data setting, increasing mAP and R1 by $11.1\%$ and $2.6\%$ points, respectively. Further enlarging the pre-training dataset to Autobot4M leads to the best performance, with VehicleMAE-V2 significantly outperforming general pre-trained models as well as other vehicle-domain methods, thereby validating its effectiveness for vehicle identity discrimination.

\begin{figure*}
\centering
\includegraphics[width=\textwidth]{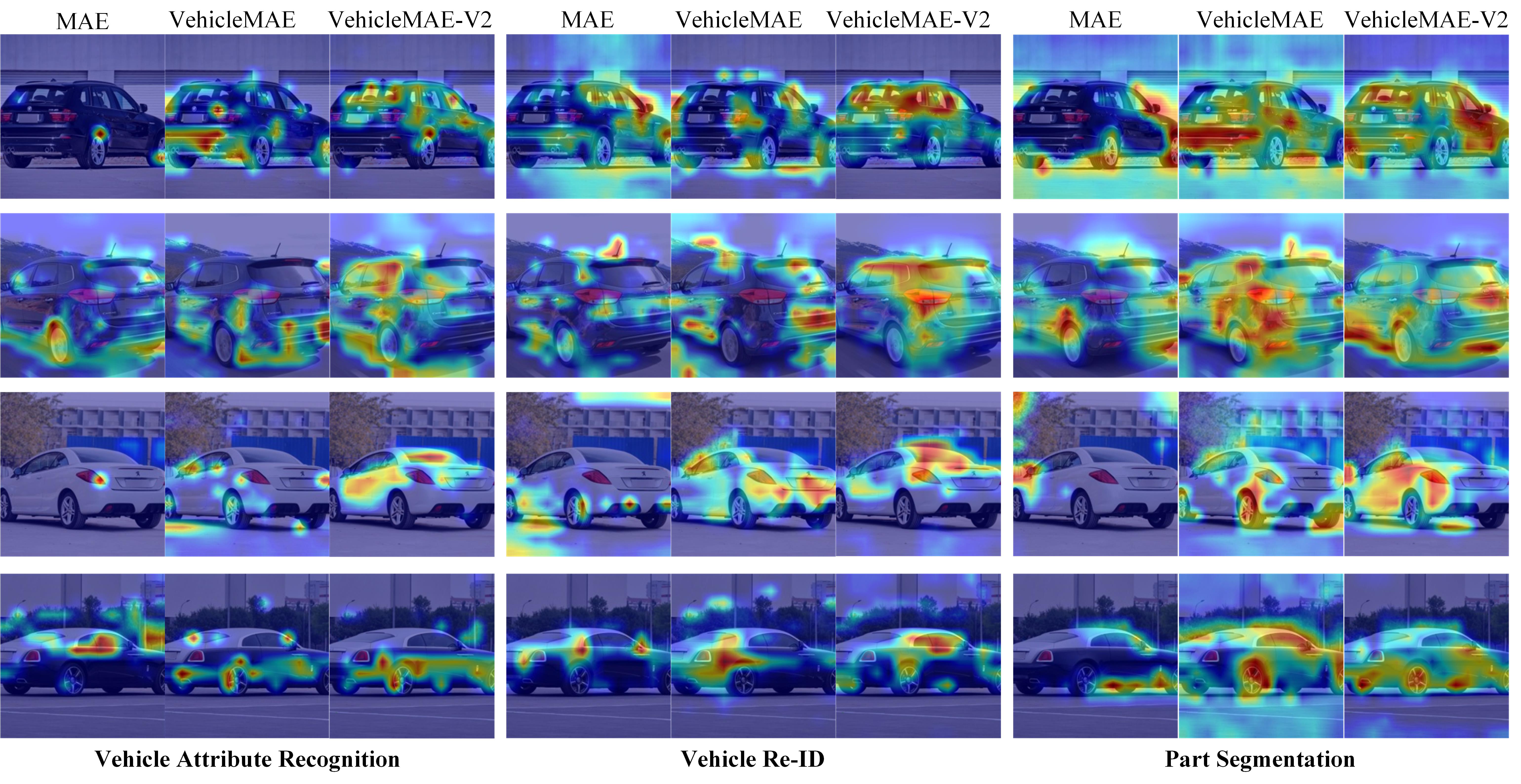}
\caption{Visualization of attention maps on different downstream tasks.}
\label{fig:attentionmaps_1} 
\end{figure*}

\noindent $\bullet$ \textbf{Vehicle Fine-grained Recognition.}
For the vehicle fine-grained recognition task, we fine-tune models based on the Transfg~\cite{he2022transfg} framework to evaluate the capability of the proposed method in modeling subtle inter-class differences. As indicated in Table~\ref{table1}, general-purpose pre-trained models exhibit limited adaptability to this task; for instance, fine-tuning DINO results in unstable training behavior. In contrast, the vehicle-domain pre-trained models VehicleMAE and VehicleMAE-V2 consistently outperform task-specific models, which may be attributed to the image–text alignment performed by the Semantics-guided Representation Module during the pre-training stage. Among them, VehicleMAE-V2 pre-trained on the Autobot4M dataset achieves the best performance, with classification accuracy (Acc) reaching $94.9\%$. It should be noted that the overall performance on this dataset is already close to saturation, leading to relatively marginal performance gains across different methods.

\noindent $\bullet$ \textbf{Vehicle Part Segmentation.}
For the vehicle part segmentation task, we further evaluate the effectiveness of the proposed pre-training strategy in enhancing pixel-level semantic understanding. Fine-tuning is conducted using the SETR~\cite{zheng2021rethinking} segmentation framework. As shown in Table~\ref{table1}, the VehicleMAE model pre-trained on Autobot1M achieves $73.29\%$ mIoU and $80.22\%$ mAcc. With the adoption of the VehicleMAE-V2 pre-training strategy, these metrics are improved to $74.66\%$ and $80.67\%$, respectively. When further scaling the pre-training dataset to Autobot4M, the model attains the best segmentation performance, achieving $75.04\%$ mIoU and $81.04\%$ mAcc, significantly outperforming general-purpose pre-trained models and other vehicle part segmentation approaches.

Overall, the experimental results across five representative downstream tasks demonstrate that VehicleMAE-V2 consistently delivers stable and notable performance improvements, spanning classification, detection, re-identification, and pixel-level segmentation tasks. These results convincingly validate the generality and effectiveness of the proposed vehicle-centric pre-training strategy across diverse application scenarios.

\begin{figure}
\centering
\includegraphics[width=3.3in]{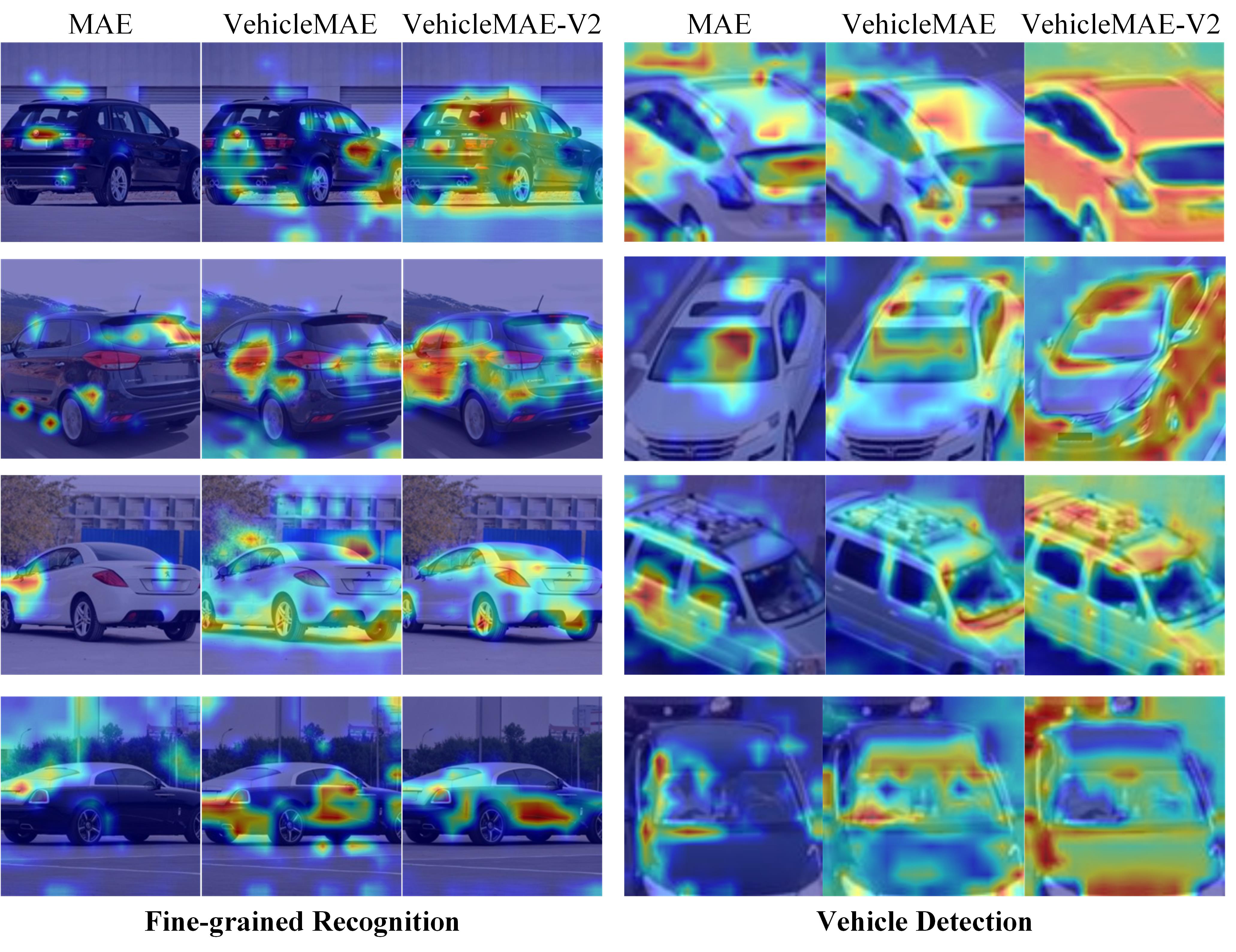}
\caption{Visualization of attention maps on different downstream tasks.}
\label{fig:attentionmaps_2} 
\end{figure}

\begin{table*}[!htbp] 
\centering 
\caption{Results of training with only $20\%$ data in downstream tasks.}  
\label{AblationStudy2}  
\footnotesize
\resizebox{1\textwidth}{!}{
\begin{tabular}{c|c|ccccc|ccc|cc|c|cc} 
\hline 
\multirow{2}{*}{\raggedright \textbf{Method}} 
&\multirow{2}{*}{\raggedright \textbf{Dataset}} 
&\multicolumn{5}{c|}{\textbf{VAR}} 
&\multicolumn{3}{c|}{\textbf{V-Det}}
&\multicolumn{2}{c|}{\textbf{V-ReID}} 
&\textbf{VFR}
&\multicolumn{2}{c}{\textbf{VPS}} \\
& &$mA$ &$Acc$ & $Prec$ & $Recall$ & $F_1$ 
& $AP_{[0.5:0.95]}$  & $AP_{0.5}$ &  $AP_{0.75}$ 
& $mAP$ & $R1$ 
& $Acc$ 
& $mIou$ & $mAcc$   \\ 
\hline 
Scratch &- &80.94 &71.33 &76.18 &79.64 &77.27   &27.7 &44.6 &31.6   &25.2 &34.9   &7.1   &39.87 &49.50 \\
MAE &ImageNet-1K &89.32 &92.65 &94.35 &94.87 &94.41   &34.4 &55.3 &37.7   &64.8 &89.7   &42.5   &64.86 &72.20 \\
MAE &Autobot1M &89.58 &92.36 &94.09 &95.06 &94.29   &34.5 &54.6 &38.0   &60.0 &85.5   &66.5   &65.04 &70.81 \\
VehicleMAE &Autobot1M &91.50 &94.53 &95.74 &96.33 &95.91   &36.3 &55.7 &39.4   &80.9 &95.2   &83.6   &68.72 &76.02 \\
VehicleMAE-V2 &Autobot1M &\textbf{92.03} &\textbf{95.10} &\textbf{96.27} &\textbf{96.64} &\textbf{96.32}   &\textbf{38.1} &\textbf{57.3} &\textbf{40.8}   &\textbf{82.1} &\textbf{95.8}   &\textbf{84.7}   &\textbf{70.30} &\textbf{76.30} \\
\hline
\end{tabular}}
\end{table*}

\begin{table*}[!htbp] 
\centering 
\caption{Ablation study on the scale of pre-training data.}  
\label{AblationStudy3}  
\footnotesize
\resizebox{1\textwidth}{!}{
\begin{tabular}{c|c|ccccc|ccc|cc|c|cc} 
\hline 
\multirow{2}{*}{\raggedright \textbf{M.R.}} 
&\multirow{2}{*}{\raggedright \textbf{Dataset}} 
&\multicolumn{5}{c|}{\textbf{VAR}} 
&\multicolumn{3}{c|}{\textbf{V-Det}}
&\multicolumn{2}{c|}{\textbf{V-ReID}} 
&\textbf{VFR}
&\multicolumn{2}{c}{\textbf{VPS}} \\
& &$mA$ &$Acc$ & $Prec$ & $Recall$ & $F_1$ 
& $AP_{[0.5:0.95]}$  & $AP_{0.5}$ &  $AP_{0.75}$ 
& $mAP$ & $R1$ 
& $Acc$ 
& $mIou$ & $mAcc$   \\ 
\hline 
\multirow{2}{*}{\raggedright $85\%$} 
&Autobot1M &92.21 &94.91 &96.00 &96.50 &96.17   &46.3 &64.6 &52.3    &83.6 &97.2   &94.3   &71.86 &78.14 \\
&Autobot4M &92.88 &95.93 &96.89 &97.14 &96.94   &46.6 &65.4 &52.6    &85.3 &97.4   &94.5   &72.95 &80.13 \\
\hline
\multirow{2}{*}{\raggedright $75\%$} 
&Autobot1M &92.43 &95.67 &96.56 &97.00 &96.70   &47.3 &67.2 &52.9   &86.6 &98.0   &94.8   &74.66 &80.67 \\
&Autobot4M &\textbf{93.19} &\textbf{95.91} &\textbf{96.89} &\textbf{97.28} &\textbf{96.98}    &\textbf{48.5} &\textbf{67.4} &\textbf{53.8}   &\textbf{87.3} &\textbf{98.0}   &\textbf{94.9}    &\textbf{75.04} &\textbf{81.04} \\
\hline
\end{tabular}}
\end{table*}

\begin{table*}[!htbp] 
\centering 
\caption{Ablation Study on Prompt Formulations in the Semantics-guided Representation Module.}
\label{AblationStudy4}  
\footnotesize
\resizebox{1\textwidth}{!}{
\begin{tabular}{c|ccccc|ccc|cc|c|cc} 
\hline 
\multirow{2}{*}{\raggedright \textbf{Method}} 
&\multicolumn{5}{c|}{\textbf{VAR}} 
&\multicolumn{3}{c|}{\textbf{V-Det}}
&\multicolumn{2}{c|}{\textbf{V-ReID}} 
&\textbf{VFR}
&\multicolumn{2}{c}{\textbf{VPS}} \\
&$mA$ &$Acc$ & $Prec$ & $Recall$ & $F_1$ 
& $AP_{[0.5:0.95]}$  & $AP_{0.5}$ &  $AP_{0.75}$ 
& $mAP$ & $R1$ 
& $Acc$ 
& $mIou$ & $mAcc$   \\ 
\hline
 
Template-free Prompt   &91.98 & 94.90& 95.96& 96.49& 96.14&   45.8& 64.9& 51.2&   85.0& 97.5&   94.5&   73.19& 79.02 \\
Randomized-template Prompt   &\textbf{92.58} &95.51 &96.38 &96.93 &96.59&   \textbf{47.3}& \textbf{67.5}& \textbf{53.4}&   86.5& 97.9&   94.8&   \textbf{74.81}& 80.47 \\
Fixed-template Prompt &92.43 &\textbf{95.67} &\textbf{96.56} &\textbf{97.00} &\textbf{96.70}    &47.3 &67.2 &52.9   &\textbf{86.6} &\textbf{98.0}   &\textbf{94.8}    &74.66 &\textbf{80.67} \\
\hline
\end{tabular}}
\end{table*}

\subsection{Ablation Study} \label{ablationStudy}

\noindent $\bullet$ \textbf{Component Analysis.}
This paper systematically analyzes the effectiveness of the proposed Contour-guided Representation Module (CRM), Semantics-guided Representation Module (SRM), and Symmetry-guided Mask Module (SMM) across five vehicle perception downstream tasks. As shown in Table~\ref{AblationStudy1}, the baseline MAE model achieves $75.5\%$ mAP and $95.4\%$ R1 on the vehicle re-identification task. After introducing CRM to guide the reconstruction of spatial image structures, the mAP and R1 improve by $7.9\%$ and $1.2\%$, respectively. Further incorporating SRM leads to additional gains of $2.7\%$ and $1.3\%$ in these two metrics. When the proposed SMM replaces the random masking strategy in MAE, the performance on the vehicle re-identification task further improves to $86.6\%$ mAP and $98.0\%$ R1. From the ablation results across the five downstream tasks, it is observed that SRM consistently brings larger performance gains than CRM, which is likely because SRM introduces the CLIP model during pre-training, enabling the model to learn richer and more discriminative semantic knowledge. In addition, under a high masking ratio ($85\%$), SMM yields more significant improvements than under a low masking ratio ($75\%$). We attribute this to the fact that, at higher masking ratios, the selection of unmasked image patches becomes more critical for representation learning, thereby highlighting the advantage of the symmetry-guided masking strategy.

\noindent $\bullet$ \textbf{Loss Analysis.}
To investigate the impact of different loss functions within each module on performance, we conducted ablation studies on the five loss functions in the Contour-guided Representation Module and Semantics-guided Representation Module, as shown in Table~\ref{AblationStudy1}. Taking the vehicle re-identification task as an example, when introducing the $L_{mim}$ loss to baseline model, the results were $79.7\%$ and $96.1\%$. After adding the $L_{cls}$ loss, the performance improved to $83.4\%$ and $96.6\%$. Building upon this, we further incorporated the consistency losses $L_{cs}$ and $L_{cf}$ from the Semantics-guided Representation Module, which boosted the results to $85.6\%$ and $97.9\%$. Finally, with the introduction of the vision-text contrastive loss $L_{vt}$ , the performance reached $86.1\%$ and $97.9\%$. Similar conclusions were observed in other tasks, with experimental results fully demonstrating the effectiveness of our proposed loss functions. 
In addition, we assign appropriate weighting coefficients to different loss terms in order to balance their numerical scales and ensure the stability of the pre-training process. As shown in the table~\ref{AblationStudy5}, we systematically analyze the impact of different loss weight configurations on downstream task performance. When all loss weights are set to 1, the model exhibits a significant performance degradation across all downstream tasks. Through further analysis of the training logs, we observe that this degradation is caused by the excessively large values of $L_{\text{mim}}$ and $L_{\text{cls}}$, which dominate the optimization process and hinder effective representation learning. To address this issue, we set both $\lambda_{\text{mim}}$ and $\lambda_{\text{cls}}$ to 0.02, which effectively suppresses their influence, achieves a more balanced contribution among different loss terms, and leads to substantial performance improvements on all downstream tasks. However, under this configuration, we further observe that the relative weight of the main loss term ($L_{\text{r}}$) becomes insufficient, causing auxiliary losses to dominate the later stages of training and potentially limiting continuous optimization toward the core reconstruction objective. Therefore, we further increase the weight of the reconstruction loss to $\lambda_{r} = 4$, thereby reinforcing the dominant role of the main loss in the overall optimization. This adjustment results in additional performance gains across multiple downstream tasks. It is worth noting that, due to the high computational cost of the pre-training process, we do not conduct a more extensive or fine-grained search over loss weight configurations.

\noindent $\bullet$ \textbf{Impact of masked token ratio on downstream tasks.}
We test four different masked ratios of 0.25, 0.50, 0.75, and 0.85 on VehicleMAE, respectively. As shown in Table~\ref{AblationStudy4}, among the five downstream tasks, the best performance is achieved when the masked ratio is set to 0.75, and its computational efficiency outperforms that of 0.25 and 0.50. With the increasing size of the pre-training dataset, computational efficiency becomes particularly important. Therefore, for VehicleMAE-V2, we train with two masked ratios: 0.75 and 0.85. Notably, in VehicleMAE-V2, while the 0.75 masked ratio still yields the best results, it achieves significant improvements over VehicleMAE at higher masked ratios. For example, in the attribute recognition task, the $mA$ metric is $92.43\%$ when the masked ratio is 0.75, and it increases to $93.01\%$ when the masked ratio is changed to 0.85. For other tasks, the results with a masked ratio of 0.85 are also comparable to those with 0.75. These experimental results and comparisons verify that VehicleMAE-V2 can adapt to higher masked ratios.

\begin{figure}
\centering
\includegraphics[width=3.3in]{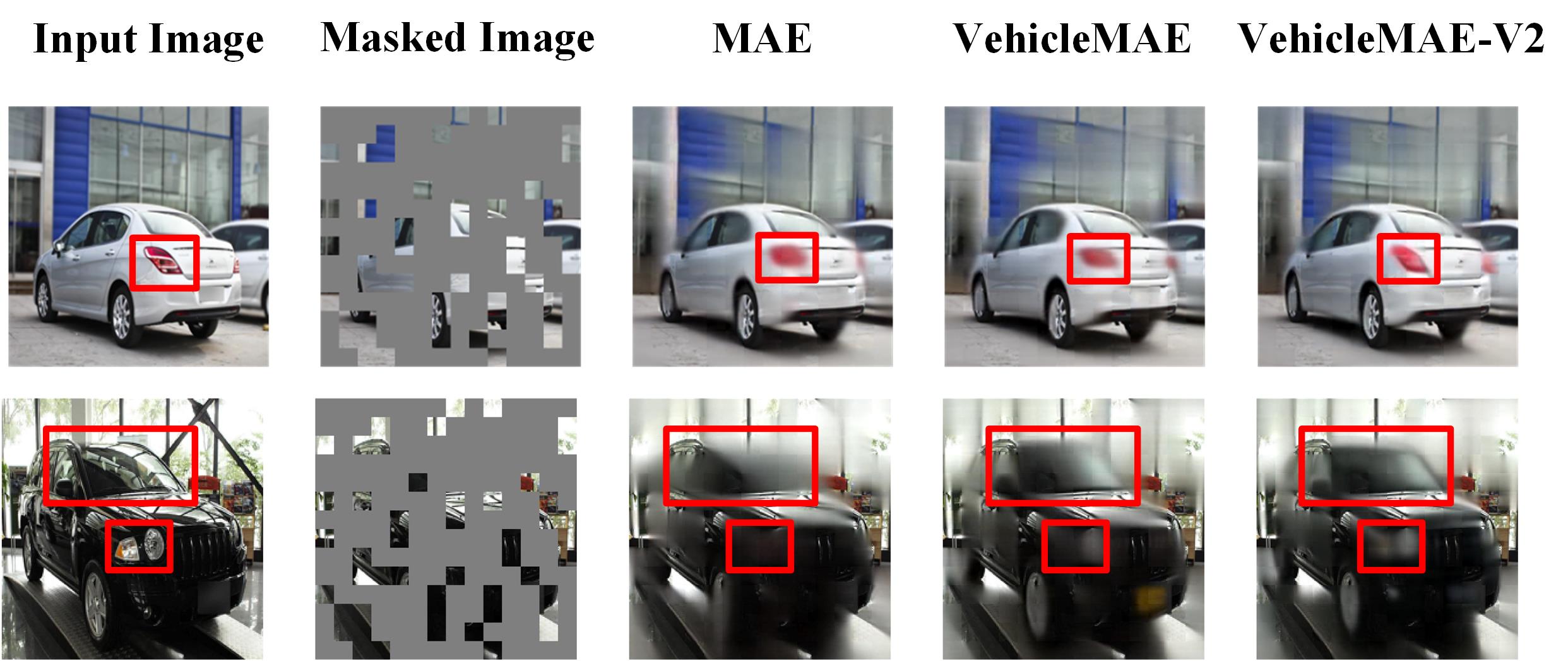}
\caption{Visualization of reconstructed vehicle images using MAE, VehicleMAE, and our newly proposed VehicleMAE-V2.}
\label{fig:reconst} 
\end{figure}

\begin{table}
\centering 
\caption{Comparison of FLOPs, MACs, and Parameters of ours and other pre-trained big models.}  
\label{efficiencyMetric}  
\footnotesize
\begin{tabular}{l|c|c|c}
\hline
\textbf{Model}    &\textbf{FLOPs (G)} &\textbf{MACs (OPs)} &\textbf{Params (M)}  \\ 
\hline 
\textbf{MoCov3}  &18.00 &17.58 &86.57   \\
\textbf{DINO}  &16.89 &16.88 &108.87   \\
\textbf{IBOT}  &18.52 &18.51 &96.29   \\
\textbf{MAE}  &9.43 &9.43 &111.65   \\
\textbf{VehicleMAE}  &10.98 &10.98 &121.62   \\
\textbf{VehicleMAE-V2}  &10.98 &10.98 &122.41   \\
\hline
\end{tabular}
\end{table}

\noindent $\bullet$ \textbf{Comparative experiments with different backbone networks.}
In this paper, we further train a pre-trained model with a larger backbone network, namely the ViT-Large/16 version, on the Autobot1M dataset, and replace the original CLIP model with the more powerful ViT-Large/14. As shown in Table~\ref{AblationStudy5}, the ViT-Large version of the pre-trained model achieves stable and consistent performance improvements across multiple vehicle downstream tasks. Taking the vehicle re-identification task as an example, the mAP metric increases from $86.6\%$ with the ViT-Base version to $89.2\%$, demonstrating a clear performance gain. Similar improvement trends are also observed for the other downstream tasks and evaluation metrics.

\noindent $\bullet$ \textbf{Impact of training data volume on downstream tasks.}
To verify the robustness of our pre-trained model, we test it with few samples in downstream tasks. Specifically, we randomly select $20\%$ of the data from the original training sets of the five downstream tasks to construct a new training set while the test set remains unchanged. As shown in Table~\ref{AblationStudy2}, we observe that when using only $20\%$ of the data for training, the fine-grained vehicle recognition task shows an initial result of $7.1\%$ without the pre-trained model. Using the MAE pre-trained model trained on ImageNet-1K improves the result to $42.5\%$, but compared to the result under the complete training set, it decreases by $48.7\%$. When using the Autobot1M dataset with the MAE pre-trained model, the result improves to $66.5\%$. With the VehicleMAE framework and the Autobot1M dataset, the result further increases to $83.6\%$. Finally, using the VehicleMAE-V2 framework proposed in this paper, the result is further improved to $84.7\%$. These experiments fully verify that our model can effectively improve the performance of the few-shot vehicle task. 

\noindent $\bullet$ \textbf{Impact of pre-training data volume on downstream tasks.}
In this paper, to further enhance the model's performance and robustness, we expand the original Autobot1M vehicle pre-training dataset, increasing the data volume to 4 million. As shown in Table~\ref{AblationStudy3}, we compare the performance of pre-trained models with different data volumes on five downstream tasks under two masked ratios of $75\%$ and $85\%$. The results indicate that pre-training on a richer pre-training dataset can effectively improve the performance on downstream tasks. Among the evaluated tasks, vehicle re-identification, attribute recognition, and part segmentation exhibit more pronounced improvements. Taking the vehicle re-identification task as an example, under the $75\%$ masking ratio, the model pre-trained on Autobot4M improves the mAP from $86.6\%$ to $87.3\%$, while under the $85\%$ masking ratio, the mAP further increases from $83.6\%$ to $85.3\%$.

\noindent $\bullet$ \textbf{Impact of prompt formulations in the Semantics-guided Representation Module on downstream tasks.}
In the Semantics-guided Representation Module (SRM), we generate prompt texts using the Bounding Box and Angle information of targets predicted by the YAEN model, thereby introducing explicit semantic priors to enhance the model’s semantic representation capability. As shown in the table~\ref{AblationStudy4}, we systematically analyze the impact of different prompt formulations on the performance of multiple vehicle downstream tasks, including Template-free Prompt, Randomized-template Prompt, and Fixed-template Prompt. The experimental results show that Template-free Prompt yields relatively inferior performance across all tasks, indicating that constructing prompt texts solely based on raw predicted parameters is insufficient to fully exploit the potential of the semantic guidance module. When template-based prompts are introduced, the model achieves significant performance improvements on all downstream tasks, which clearly demonstrates the positive role of structured semantic information in facilitating cross-modal alignment and discriminative representation learning. On this basis, we further explore Randomized-template Prompt by incorporating a template pool generated by a large language model to increase the diversity of prompt texts, aiming to further enhance the generalization ability of SRM. However, the experimental results indicate that Randomized-template Prompt exhibits performance comparable to, and in some metrics slightly inferior to, the Fixed-template Prompt. We attribute this to the semantic perturbations introduced by random templates, which may partially weaken semantic consistency across samples and consequently affect the modeling of stable semantic structures. Based on the above analysis, we finally adopt Fixed-template Prompt as the default configuration of SRM, as it achieves a better balance between semantic consistency and overall performance.

\noindent $\bullet$ \textbf{Impact of pre-training epochs on downstream tasks.}
We plot the performance variation curves of five downstream tasks under different pre-training epochs. Specifically, we select one model every 10 epochs, resulting in 10 pre-trained models from 10 to 100 epochs for evaluation. All models are based on the VehicleMAE-V2 framework and are pre-trained on the Autobot1M and Autobot4M datasets. As shown in Fig.~\ref{fig:epoch} and Fig.~\ref{fig:epoch_det}, the optimal pre-training epochs vary significantly across different types of downstream tasks. For example, in classification tasks such as vehicle attribute recognition and fine-grained recognition, the performance consistently improves as the number of pre-training epochs increases; in contrast, for vehicle detection tasks, the best performance typically emerges at the middle stage of pre-training, around 60 epochs on Autobot1M and 50 epochs on Autobot4M.

\subsection{Efficiency Analysis} \label{convergeAnalysis} 
As shown in Table~\ref{efficiencyMetric}, we compare the FLOPs\footnote{\url{https://pypi.org/project/ptflops/}}, MACs\footnote{\url{https://pypi.org/project/thop/}}, and Parameters of the VehicleMAE-V2 model with other pre-trained large models. Notably, the FLOPs and MACs of VehicleMAE-V2 are both 10.98G, which is the same as those of VehicleMAE. In terms of the number of parameters, VehicleMAE has 121.62M, while VehicleMAE-V2 has 122.41M. From the comparison of these efficiency metrics, we find that VehicleMAE-V2 increases only slightly in the number of parameters compared to VehicleMAE. Meanwhile, VehicleMAE-V2 outperforms VehicleMAE across five downstream vehicle tasks.

\subsection{Visualization} \label{visualization} 
In this section, we visualize the reconstructions of feature maps, masked images, and experimental results for downstream tasks on MAE, VehicleMAE, and our newly proposed large model, VehicleMAE-V2.  As shown in Figure~\ref{fig:attentionmaps_1} and Figure~\ref{fig:attentionmaps_2}, we provide feature maps of vehicle images for five downstream tasks, with the visualized feature maps for proposals in the vehicle detection task. Specifically, we utilize GradCAM\footnote{\url{https://mmpretrain.readthedocs.io/en/latest/useful_tools/cam_visualization.html}} to visualize the feature maps from the 11th Transformer block. Compared to MAE and VehicleMAE, our newly proposed VehicleMAE-V2 exhibits higher responses in critical areas. Consequently, our model performs better in multiple vehicle perception tasks. As shown in Figure~\ref{fig:reconst}, we present the reconstruction results of masked image patches by the three models. Compared to MAE, VehicleMAE reconstructs the overall structure of vehicles better, but the reconstruction in some detailed areas is not clear. From the reconstructed results, our newly proposed VehicleMAE-V2 is capable of clearly and accurately reconstructing detailed areas of vehicles, such as headlights and roofs. In the images, we annotate these areas with red bounding boxes.

\section{Limitations and Future Work}  
\label{sec::Discussion} 
Although the proposed VehicleMAE achieves notable performance improvements across multiple vehicle perception tasks, several aspects still merit further investigation and refinement. The current pre-training model adopts a vehicle-centric design, focusing primarily on the structural and semantic representations of individual vehicle instances. As a result, by simply loading the pre-trained parameters, the model transfers effectively to vehicle-oriented tasks such as recognition and segmentation. However, in large-scale scene perception tasks such as vehicle detection, where vehicles typically appear as local targets embedded in complex backgrounds, directly applying the pre-trained model often yields sub-optimal performance and requires additional task-specific detection frameworks for effective adaptation. Future research explores more unified and detection-friendly pre-training strategies, enabling the learned representations to generalize more naturally to complex scene perception tasks such as detection. Furthermore, during the pre-training stage, VehicleMAE relies solely on RGB images and textual information for visual–language joint modeling, without fully considering the diverse and heterogeneous modalities commonly present in intelligent transportation scenarios. Future work further extends the multimodal pre-training paradigm by incorporating additional modalities, such as infrared images, depth information, LiDAR point clouds, and event-based sensors, to construct a more general and robust multimodal vehicle foundation model.

\section{Conclusion}  
\label{sec6} 

In this study, we propose a novel vehicle-centric multimodal structured pre-training framework. For each input image, we first segment it into disjoint patches. Using a symmetry-guided mask module, we selectively mask the top 75$\%$ of the most salient image patches. The remaining unmasked patches are transformed into feature embeddings, which are fed into a Vision Transformer (ViT) backbone. A Transformer decoder is then employed to reconstruct the masked patches. Subsequently, we design a contour-guided representation module that obtains contour maps of the input images via an edge detector and extracts contour features using a shared Transformer encoder. By minimizing the probability distribution divergence between contour features and reconstructed features, this module guides the reconstruction of the vehicle’s spatial structure at the pixel level. Finally, we develop a semantics-guided representation module that enhances the model’s semantic understanding through image-text contrastive learning, while also leveraging cross-modal knowledge distillation based on the pre-trained CLIP model to learn rich vehicle-related knowledge. This approach mitigates feature confusion caused by insufficient semantic understanding during masked reconstruction. To address the data disparity, we have developed a large-scale vehicle pre-training dataset named Autobot4M, which comprises approximately 4 million vehicle images and 12,693 textual descriptions. The pre-trained model is assessed on five vehicle perception downstream tasks, including vehicle attribute recognition, re-identification, fine-grained recognition, object detection, and part segmentation. Comprehensive experiments thoroughly demonstrate the efficacy and benefits of our proposed VehicleMAE-V2 framework.


\bibliographystyle{IEEEtran}
\bibliography{reference}

\begin{thebibliography}{10}
\providecommand{\url}[1]{#1}
\csname url@samestyle\endcsname
\providecommand{\newblock}{\relax}
\providecommand{\bibinfo}[2]{#2}
\providecommand{\BIBentrySTDinterwordspacing}{\spaceskip=0pt\relax}
\providecommand{\BIBentryALTinterwordstretchfactor}{4}
\providecommand{\BIBentryALTinterwordspacing}{\spaceskip=\fontdimen2\font plus
\BIBentryALTinterwordstretchfactor\fontdimen3\font minus
  \fontdimen4\font\relax}
\providecommand{\BIBforeignlanguage}[2]{{%
\expandafter\ifx\csname l@#1\endcsname\relax
\typeout{** WARNING: IEEEtran.bst: No hyphenation pattern has been}%
\typeout{** loaded for the language `#1'. Using the pattern for}%
\typeout{** the default language instead.}%
\else
\language=\csname l@#1\endcsname
\fi
#2}}
\providecommand{\BIBdecl}{\relax}
\BIBdecl

\bibitem{yu2022embedding}
Y.~Yu, H.~Liu, Y.~Fu, W.~Jia, J.~Yu, and Z.~Yan, ``Embedding pose information
  for multiview vehicle model recognition,'' \emph{IEEE Transactions on
  Circuits and Systems for Video Technology}, vol.~32, no.~8, pp. 5467--5480,
  2022.

\bibitem{liang2024scene}
Y.~Liang, G.~Shi, and J.~Wu, ``Scene prior constrained self-paced learning for
  unsupervised satellite video vehicle detection,'' \emph{IEEE Transactions on
  Circuits and Systems for Video Technology}, 2024.

\bibitem{wang2024class}
J.~Wang, T.~Dai, X.~Zhao, {\'A}.~F. Garc{\'\i}a-Fern{\'a}ndez, E.~G. Lim, and
  J.~Xiao, ``Class activation map calibration for weakly supervised semantic
  segmentation,'' \emph{IEEE Transactions on Circuits and Systems for Video
  Technology}, 2024.

\bibitem{ran2025context}
Z.~Ran, Z.~Xiao, X.~Lu, X.~Wei, and W.~Liu, ``Context-aided semantic-aware
  self-alignment for video-based person re-identification,'' \emph{IEEE
  Transactions on Circuits and Systems for Video Technology}, 2025.

\bibitem{wang2024pedestrian}
X.~Wang, J.~Jin, C.~Li, J.~Tang, C.~Zhang, and W.~Wang, ``Pedestrian attribute
  recognition via clip based prompt vision-language fusion,'' \emph{IEEE
  Transactions on Circuits and Systems for Video Technology}, 2024.

\bibitem{he2016deep}
K.~He, X.~Zhang, S.~Ren, and J.~Sun, ``Deep residual learning for image
  recognition,'' in \emph{Proceedings of the IEEE conference on computer vision
  and pattern recognition}, 2016, pp. 770--778.

\bibitem{VIT}
A.~Dosovitskiy, L.~Beyer, A.~Kolesnikov, D.~Weissenborn, X.~Zhai,
  T.~Unterthiner, M.~Dehghani, M.~Minderer, G.~Heigold, S.~Gelly \emph{et~al.},
  ``An image is worth 16x16 words: Transformers for image recognition at
  scale,'' in \emph{International Conference on Learning Representations},
  2021.

\bibitem{yuan2024surveillance}
T.~Yuan, X.~Zhang, B.~Liu, K.~Liu, J.~Jin, and Z.~Jiao, ``Surveillance
  video-and-language understanding: from small to large multimodal models,''
  \emph{IEEE Transactions on Circuits and Systems for Video Technology}, 2024.

\bibitem{chen2023beyond}
W.~Chen, X.~Xu, J.~Jia, H.~Luo, Y.~Wang, F.~Wang, R.~Jin, and X.~Sun, ``Beyond
  appearance: a semantic controllable self-supervised learning framework for
  human-centric visual tasks,'' in \emph{Proceedings of the IEEE/CVF conference
  on computer vision and pattern recognition}, 2023, pp. 15\,050--15\,061.

\bibitem{radford2018GPT1}
A.~Radford, K.~Narasimhan, T.~Salimans, I.~Sutskever \emph{et~al.}, ``Improving
  language understanding by generative pre-training,'' \emph{OpenAI blog},
  2018.

\bibitem{radford2019GPT2}
A.~Radford, J.~Wu, R.~Child, D.~Luan, D.~Amodei, I.~Sutskever \emph{et~al.},
  ``Language models are unsupervised multitask learners,'' \emph{OpenAI blog},
  vol.~1, no.~8, p.~9, 2019.

\bibitem{brown2020GPT3}
T.~Brown, B.~Mann, N.~Ryder, M.~Subbiah, J.~D. Kaplan, P.~Dhariwal,
  A.~Neelakantan, P.~Shyam, G.~Sastry, A.~Askell \emph{et~al.}, ``Language
  models are few-shot learners,'' \emph{Advances in neural information
  processing systems}, vol.~33, pp. 1877--1901, 2020.

\bibitem{openai2023gpt4}
OpenAI, ``Gpt-4 technical report,'' 2023.

\bibitem{touvron2023llama}
H.~Touvron, T.~Lavril, G.~Izacard, X.~Martinet, M.-A. Lachaux, T.~Lacroix,
  B.~Rozi{\`e}re, N.~Goyal, E.~Hambro, F.~Azhar \emph{et~al.}, ``Llama: Open
  and efficient foundation language models,'' \emph{arXiv preprint
  arXiv:2302.13971}, 2023.

\bibitem{liu2021swintransformer}
Z.~Liu, Y.~Lin, Y.~Cao, H.~Hu, Y.~Wei, Z.~Zhang, S.~Lin, and B.~Guo, ``Swin
  transformer: Hierarchical vision transformer using shifted windows,'' in
  \emph{Proceedings of the IEEE/CVF International Conference on Computer
  Vision}, 2021, pp. 10\,012--10\,022.

\bibitem{li2023blip}
J.~Li, D.~Li, S.~Savarese, and S.~Hoi, ``Blip-2: Bootstrapping language-image
  pre-training with frozen image encoders and large language models,'' in
  \emph{International Conference on Machine Learning}, 2023, pp.
  19\,730--19\,742.

\bibitem{xue2024ulip}
L.~Xue, N.~Yu, S.~Zhang, A.~Panagopoulou, J.~Li, R.~Mart{\'\i}n-Mart{\'\i}n,
  J.~Wu, C.~Xiong, R.~Xu, J.~C. Niebles \emph{et~al.}, ``Ulip-2: Towards
  scalable multimodal pre-training for 3d understanding,'' in \emph{Proceedings
  of the IEEE/CVF Conference on Computer Vision and Pattern Recognition}, 2024,
  pp. 27\,091--27\,101.

\bibitem{he2022mae}
K.~He, X.~Chen, S.~Xie, Y.~Li, P.~Doll{\'a}r, and R.~Girshick, ``Masked
  autoencoders are scalable vision learners,'' in \emph{Proceedings of the
  IEEE/CVF Conference on Computer Vision and Pattern Recognition}, 2022, pp.
  16\,000--16\,009.

\bibitem{radford2021learning}
A.~Radford, J.~W. Kim, C.~Hallacy, A.~Ramesh, G.~Goh, S.~Agarwal, G.~Sastry,
  A.~Askell, P.~Mishkin, J.~Clark \emph{et~al.}, ``Learning transferable visual
  models from natural language supervision,'' in \emph{International Conference
  on Machine Learning}, 2021, pp. 8748--8763.

\bibitem{wang2024structural}
X.~Wang, W.~Wu, C.~Li, Z.~Zhao, Z.~Chen, Y.~Shi, and J.~Tang, ``Structural
  information guided multimodal pre-training for vehicle-centric perception,''
  in \emph{Proceedings of the AAAI Conference on Artificial Intelligence},
  vol.~38, no.~6, 2024, pp. 5624--5632.

\bibitem{wu2024vfm}
W.~Wu, F.~Hong, X.~Wang, C.~Li, and J.~Tang, ``Vfm-det: Towards
  high-performance vehicle detection via large foundation models,'' \emph{arXiv
  preprint arXiv:2408.13031}, 2024.

\bibitem{he2020momentum}
K.~He, H.~Fan, Y.~Wu, S.~Xie, and R.~Girshick, ``Momentum contrast for
  unsupervised visual representation learning,'' in \emph{Proceedings of the
  IEEE/CVF conference on computer vision and pattern recognition}, 2020, pp.
  9729--9738.

\bibitem{chen2020simple}
T.~Chen, S.~Kornblith, M.~Norouzi, and G.~Hinton, ``A simple framework for
  contrastive learning of visual representations,'' in \emph{International
  Conference on Machine Learning}, 2020, pp. 1597--1607.

\bibitem{qiu2024camera}
M.~Qiu, Y.~Lu, X.~Li, and Q.~Lu, ``Camera-aware differentiated clustering with
  focal contrastive learning for unsupervised vehicle re-identification,''
  \emph{IEEE Transactions on Circuits and Systems for Video Technology},
  vol.~34, no.~10, pp. 10\,121--10\,134, 2024.

\bibitem{wang2025contrastive}
J.~Wang, X.~Li, X.~Dai, S.~Zhuang, and M.~Qi, ``Contrastive learning-based
  joint pre-training for unsupervised domain adaptive person
  re-identification,'' \emph{Multimedia Systems}, vol.~31, no.~2, pp. 1--15,
  2025.

\bibitem{xu2025sdcluster}
H.~Xu, C.~Zhang, P.~Yue, and K.~Wang, ``Sdcluster: A clustering based
  self-supervised pre-training method for semantic segmentation of remote
  sensing images,'' \emph{ISPRS Journal of Photogrammetry and Remote Sensing},
  vol. 223, pp. 1--14, 2025.

\bibitem{kenton2019bert}
J.~D. M.-W.~C. Kenton and L.~K. Toutanova, ``Bert: Pre-training of deep
  bidirectional transformers for language understanding,'' in \emph{Proceedings
  of NAACL-HLT}, 2019, pp. 4171--4186.

\bibitem{baobeit}
H.~Bao, L.~Dong, S.~Piao, and F.~Wei, ``Beit: Bert pre-training of image
  transformers,'' in \emph{International Conference on Learning
  Representations}, 2022.

\bibitem{zhu2025driving}
G.~Zhu, Z.~Qin, E.~Zhou, Y.~Ding, and Z.~Qin, ``Driving mutual advancement of
  3d reconstruction and inpainting for masked faces,'' \emph{Pattern
  Recognition}, vol. 158, p. 110975, 2025.

\bibitem{gong2025rethinking}
T.~Gong, Q.~Chu, B.~Liu, and N.~Yu, ``Rethinking masked data reconstruction
  pretraining for strong 3d action representation learning,'' in
  \emph{Proceedings of the AAAI Conference on Artificial Intelligence},
  vol.~39, no.~3, 2025, pp. 3149--3157.

\bibitem{ni2025maskgwm}
J.~Ni, Y.~Guo, Y.~Liu, R.~Chen, L.~Lu, and Z.~Wu, ``Maskgwm: A generalizable
  driving world model with video mask reconstruction,'' in \emph{Proceedings of
  the Computer Vision and Pattern Recognition Conference}, 2025, pp.
  22\,381--22\,391.

\bibitem{zhang2024self}
W.-L. Zhang, R.-S. Jia, H.~Wang, C.-Y. Che, and H.-M. Sun, ``A self-supervised
  learning network for student engagement recognition from facial
  expressions,'' \emph{IEEE Transactions on Circuits and Systems for Video
  Technology}, 2024.

\bibitem{li2025cross}
B.~Li, J.~Chen, G.~Li, D.~Zhang, X.~Bao, and D.~Huang, ``Cross-modal
  contrastive masked autoencoder for compressed video pre-training,''
  \emph{IEEE Transactions on Image Processing}, 2025.

\bibitem{wu2025cm3ae}
W.~Wu, X.~Wang, C.~Li, B.~Jiang, J.~Tang, B.~Luo, and Q.~Liu, ``Cm3ae: A
  unified rgb frame and event-voxel/-frame pre-training framework,'' in
  \emph{Proceedings of the 33rd ACM International Conference on Multimedia},
  2025, pp. 2159--2168.

\bibitem{lu2024cross}
M.~Lu, S.~Yang, X.~Lu, and J.~Liu, ``Cross-modal contrastive pre-training for
  few-shot skeleton action recognition,'' \emph{IEEE Transactions on Circuits
  and Systems for Video Technology}, vol.~34, no.~10, pp. 9798--9807, 2024.

\bibitem{zuo2023plip}
J.~Zuo, C.~Yu, N.~Sang, and C.~Gao, ``Plip: Language-image pre-training for
  person representation learning,'' \emph{arXiv preprint arXiv:2305.08386},
  2023.

\bibitem{wald2025revisiting}
T.~Wald, C.~Ulrich, S.~Lukyanenko, A.~Goncharov, A.~Paderno, M.~Miller,
  L.~Maerkisch, P.~Jaeger, and K.~Maier-Hein, ``Revisiting mae pre-training for
  3d medical image segmentation,'' in \emph{Proceedings of the Computer Vision
  and Pattern Recognition Conference}, 2025, pp. 5186--5196.

\bibitem{wang2024HDXrayMAE}
X.~Wang, Y.~Li, W.~Wu, J.~Jin, Y.~Rong, B.~Jiang, C.~Li, and J.~Tang,
  ``Pre-training on high definition x-ray images: An experimental study,''
  \emph{arXiv preprint arXiv:2404.17926}, 2024.

\bibitem{wang2025roma}
F.~Wang, Y.~Wang, M.~Chen, H.~Zhao, Y.~Sun, S.~Wang, H.~Wang, D.~Wang, L.~Lan,
  W.~Yang \emph{et~al.}, ``Roma: Scaling up mamba-based foundation models for
  remote sensing,'' \emph{arXiv preprint arXiv:2503.10392}, 2025.

\bibitem{hu2025rs}
H.~Hu, P.~Wang, H.~Bi, B.~Tong, Z.~Wang, W.~Diao, H.~Chang, Y.~Feng, Z.~Zhang,
  Y.~Wang \emph{et~al.}, ``Rs-vheat: Heat conduction guided efficient remote
  sensing foundation model,'' in \emph{Proceedings of the IEEE/CVF
  International Conference on Computer Vision}, 2025, pp. 9876--9887.

\bibitem{zhudeformable}
X.~Zhu, W.~Su, L.~Lu, B.~Li, X.~Wang, and J.~Dai, ``Deformable detr: Deformable
  transformers for end-to-end object detection,'' in \emph{International
  Conference on Learning Representations}, 2022.

\bibitem{li2022exploring}
Y.~Li, H.~Mao, R.~Girshick, and K.~He, ``Exploring plain vision transformer
  backbones for object detection,'' in \emph{European conference on computer
  vision}.\hskip 1em plus 0.5em minus 0.4em\relax Springer, 2022, pp. 280--296.

\bibitem{cheng2022simple}
X.~Cheng, M.~Jia, Q.~Wang, and J.~Zhang, ``A simple visual-textual baseline for
  pedestrian attribute recognition,'' \emph{IEEE Transactions on Circuits and
  Systems for Video Technology}, vol.~32, no.~10, pp. 6994--7004, 2022.

\bibitem{wu2024selective}
J.~Wu, Y.~Huang, M.~Gao, Y.~Niu, M.~Yang, Z.~Gao, and J.~Zhao, ``Selective and
  orthogonal feature activation for pedestrian attribute recognition,'' in
  \emph{Proceedings of the AAAI Conference on Artificial Intelligence},
  vol.~38, no.~6, 2024, pp. 6039--6047.

\bibitem{bui2024c2t}
D.~C. Bui, T.~V. Le, and B.~H. Ngo, ``C2t-net: Channel-aware cross-fused
  transformer-style networks for pedestrian attribute recognition,'' in
  \emph{Proceedings of the IEEE/CVF Winter Conference on Applications of
  Computer Vision}, 2024, pp. 351--358.

\bibitem{he2021transreid}
S.~He, H.~Luo, P.~Wang, F.~Wang, H.~Li, and W.~Jiang, ``Transreid:
  Transformer-based object re-identification,'' in \emph{Proceedings of the
  IEEE/CVF International Conference on Computer Vision}, 2021, pp.
  15\,013--15\,022.

\bibitem{li2024day}
H.~Li, J.~Chen, A.~Zheng, Y.~Wu, and Y.~Luo, ``Day-night cross-domain vehicle
  re-identification,'' in \emph{Proceedings of the IEEE/CVF Conference on
  Computer Vision and Pattern Recognition}, 2024, pp. 12\,626--12\,635.

\bibitem{shen2023triplet}
F.~Shen, X.~Du, L.~Zhang, X.~Shu, and J.~Tang, ``Triplet contrastive
  representation learning for unsupervised vehicle re-identification,''
  \emph{arXiv e-prints}, pp. arXiv--2301, 2023.

\bibitem{huang2022deep}
W.~Huang, W.~Li, L.~Tang, X.~Zhu, and B.~Zou, ``A deep learning framework for
  accurate vehicle yaw angle estimation from a monocular camera based on part
  arrangement,'' \emph{Sensors}, vol.~22, no.~20, p. 8027, 2022.

\bibitem{he2020bdcn}
J.~He, S.~Zhang, M.~Yang, Y.~Shan, and T.~Huang, ``Bdcn: Bi-directional cascade
  network for perceptual edge detection,'' \emph{IEEE Transactions on Pattern
  Analysis and Machine Intelligence}, vol.~44, no.~1, pp. 100--113, 2020.

\bibitem{liu2025knowledge}
F.~Liu, K.~Huang, and Q.~Li, ``Knowledge-driven multi-branch interaction
  network for vehicle re-identification,'' \emph{IEEE Transactions on
  Intelligent Transportation Systems}, 2025.

\bibitem{islam2024diffusemix}
K.~Islam, M.~Z. Zaheer, A.~Mahmood, and K.~Nandakumar, ``Diffusemix:
  Label-preserving data augmentation with diffusion models,'' in
  \emph{Proceedings of the IEEE/CVF Conference on Computer Vision and Pattern
  Recognition}, 2024, pp. 27\,621--27\,630.

\bibitem{yang2025golden}
G.~Yang, Y.~Wang, D.~Shi, and Y.~Wang, ``Golden cudgel network for real-time
  semantic segmentation,'' in \emph{Proceedings of the Computer Vision and
  Pattern Recognition Conference}, 2025, pp. 25\,367--25\,376.

\bibitem{MoCov32021}
X.~Chen, S.~Xie, and K.~He, ``An empirical study of training self-supervised
  vision transformers,'' in \emph{Proceedings of the IEEE/CVF International
  Conference on Computer Vision}, 2021, pp. 9620--9629.

\bibitem{caron2021emerging}
M.~Caron, H.~Touvron, I.~Misra, H.~J{\'e}gou, J.~Mairal, P.~Bojanowski, and
  A.~Joulin, ``Emerging properties in self-supervised vision transformers,'' in
  \emph{Proceedings of the IEEE/CVF International Conference on Computer
  Vision}, 2021, pp. 9650--9660.

\bibitem{zhouimage}
J.~Zhou, C.~Wei, H.~Wang, W.~Shen, C.~Xie, A.~Yuille, and T.~Kong, ``Image bert
  pre-training with online tokenizer,'' in \emph{International Conference on
  Learning Representations}, 2022.

\bibitem{yan2017exploiting}
K.~Yan, Y.~Tian, Y.~Wang, W.~Zeng, and T.~Huang, ``Exploiting multi-grain
  ranking constraints for precisely searching visually-similar vehicles,'' in
  \emph{Proceedings of the IEEE International Conference on Computer Vision},
  2017, pp. 562--570.

\bibitem{han2soda10m}
J.~Han, X.~Liang, H.~Xu, K.~Chen, H.~Lanqing, J.~Mao, C.~Ye, W.~Zhang, Z.~Li,
  X.~Liang \emph{et~al.}, ``Soda10m: A large-scale 2d self/semi-supervised
  object detection dataset for autonomous driving,'' in \emph{Thirty-fifth
  Conference on Neural Information Processing Systems Datasets and Benchmarks
  Track}, 2021.

\bibitem{liu2016Veri776}
X.~Liu, W.~Liu, T.~Mei, and H.~Ma, ``A deep learning-based approach to
  progressive vehicle re-identification for urban surveillance,'' in
  \emph{Computer Vision--ECCV 2016: 14th European Conference, Amsterdam, The
  Netherlands, October 11-14, 2016, Proceedings, Part II 14}.\hskip 1em plus
  0.5em minus 0.4em\relax Springer, 2016, pp. 869--884.

\bibitem{krause2013Stanfordcars}
J.~Krause, M.~Stark, J.~Deng, and L.~Fei-Fei, ``3d object representations for
  fine-grained categorization,'' in \emph{Proceedings of the IEEE International
  Conference on Computer Vision workshops}, 2013, pp. 554--561.

\bibitem{he2022partimagenet}
J.~He, S.~Yang, S.~Yang, A.~Kortylewski, X.~Yuan, J.-N. Chen, S.~Liu, C.~Yang,
  Q.~Yu, and A.~Yuille, ``Partimagenet: A large, high-quality dataset of
  parts,'' in \emph{Computer Vision--ECCV 2022: 17th European Conference, Tel
  Aviv, Israel, October 23--27, 2022, Proceedings, Part VIII}.\hskip 1em plus
  0.5em minus 0.4em\relax Springer, 2022, pp. 128--145.

\bibitem{cordts2016cityscapes}
M.~Cordts, M.~Omran, S.~Ramos, T.~Rehfeld, M.~Enzweiler, R.~Benenson,
  U.~Franke, S.~Roth, and B.~Schiele, ``The cityscapes dataset for semantic
  urban scene understanding,'' in \emph{Proceedings of the IEEE conference on
  computer vision and pattern recognition}, 2016, pp. 3213--3223.

\bibitem{adamW}
I.~Loshchilov and F.~Hutter, ``Decoupled weight decay regularization,'' in
  \emph{International Conference on Learning Representations}, 2019.

\bibitem{cheng2022VTB}
X.~Cheng, M.~Jia, Q.~Wang, and J.~Zhang, ``A simple visual-textual baseline for
  pedestrian attribute recognition,'' \emph{IEEE Transactions on Circuits and
  Systems for Video Technology}, vol.~32, no.~10, pp. 6994--7004, 2022.

\bibitem{he2022transfg}
J.~He, J.-N. Chen, S.~Liu, A.~Kortylewski, C.~Yang, Y.~Bai, and C.~Wang,
  ``Transfg: A transformer architecture for fine-grained recognition,'' in
  \emph{Proceedings of the AAAI Conference on Artificial Intelligence},
  vol.~36, no.~1, 2022, pp. 852--860.

\bibitem{zheng2021rethinking}
S.~Zheng, J.~Lu, H.~Zhao, X.~Zhu, Z.~Luo, Y.~Wang, Y.~Fu, J.~Feng, T.~Xiang,
  P.~H. Torr \emph{et~al.}, ``Rethinking semantic segmentation from a
  sequence-to-sequence perspective with transformers,'' in \emph{Proceedings of
  the IEEE/CVF conference on computer vision and pattern recognition}, 2021,
  pp. 6881--6890.

\end{thebibliography}

\end{document}